\PassOptionsToPackage{table}{xcolor}
\documentclass[10pt]{article} %
\usepackage[preprint]{tmlr}

\usepackage{pifont}        
\usepackage{amsmath, amsfonts, bm}
\usepackage{booktabs}
\usepackage{arydshln}
\usepackage{multirow}
\usepackage{subcaption}
\usepackage{enumitem}
\usepackage{mathabx}
\usepackage{bbm}
\usepackage{wrapfig}
\usepackage{float}
\usepackage{graphicx}
\usepackage{hyperref}       
\usepackage{url}
\usepackage{cleveref}

\def\eqref#1{equation~\ref{#1}}

\def\1{\bm{1}}

\DeclareMathAlphabet{\mathsfit}{\encodingdefault}{\sfdefault}{m}{sl}
\SetMathAlphabet{\mathsfit}{bold}{\encodingdefault}{\sfdefault}{bx}{n}

\title{
Scaling Image Tokenizers with Grouped Spherical Quantization
}

\author{Jiangtao Wang$^{1}$ , Zhen Qin$^{2}$, Yifan Zhang$^{3}$, Vincent Tao Hu$^{4}$, \\ Björn Ommer$^{4}$, Rania Briq$^{1}$,  Stefan Kesselheim$^{1}$ \\
\\
Jülich Supercomputing Centre$^{1}$, TapTap$^{2}$, Tsinghua University$^{3}$, \\ CompVis @ LMU Munich, MCML$^{4}$ \\ \\
\href{https://github.com/HelmholtzAI-FZJ/flex_gen}{\color{blue}{Training Code}}\ \& \href{https://huggingface.co/collections/HelmholtzAI-FZJ/grouped-spherical-quantization-674d6f9f548e472d0eaf179e}{\color{blue}{Checkpoints}} 
}

\begin{document}

\maketitle
\begin{abstract}
Vision tokenizers have gained a lot of attraction due to their scalability and compactness; previous works depend on old school GAN-based hyperparameters, biased comparisons, and a lack of comprehensive analysis of the scaling behaviours.
To tackle those issues, we introduce Grouped Spherical Quantization (GSQ), featuring spherical codebook initialization and lookup regularization to constrain codebook latent to a spherical surface. Our empirical analysis of image tokenizer training strategies demonstrates that GSQ-GAN achieves superior reconstruction quality over state-of-the-art methods with fewer training iterations, providing a solid foundation for scaling studies. Building on this, we systematically examine the scaling behaviours of GSQ—specifically in latent dimensionality, codebook size, and compression ratios—and their impact on model performance. Our findings reveal distinct behaviours at high and low spatial compression levels, underscoring challenges in representing high-dimensional latent spaces. We show that GSQ can restructure high-dimensional latent into compact, low-dimensional spaces, thus enabling efficient scaling with improved quality. As a result, GSQ-GAN achieves a 16× down-sampling with a reconstruction FID (rFID) of 0.50. %

\end{abstract}

\section{Introduction}
\label{sec:intro}

\begin{figure}[hbt!]
    \centering
    \begin{minipage}[t]{0.6\textwidth}
    \centering
    \includegraphics[width=\textwidth]{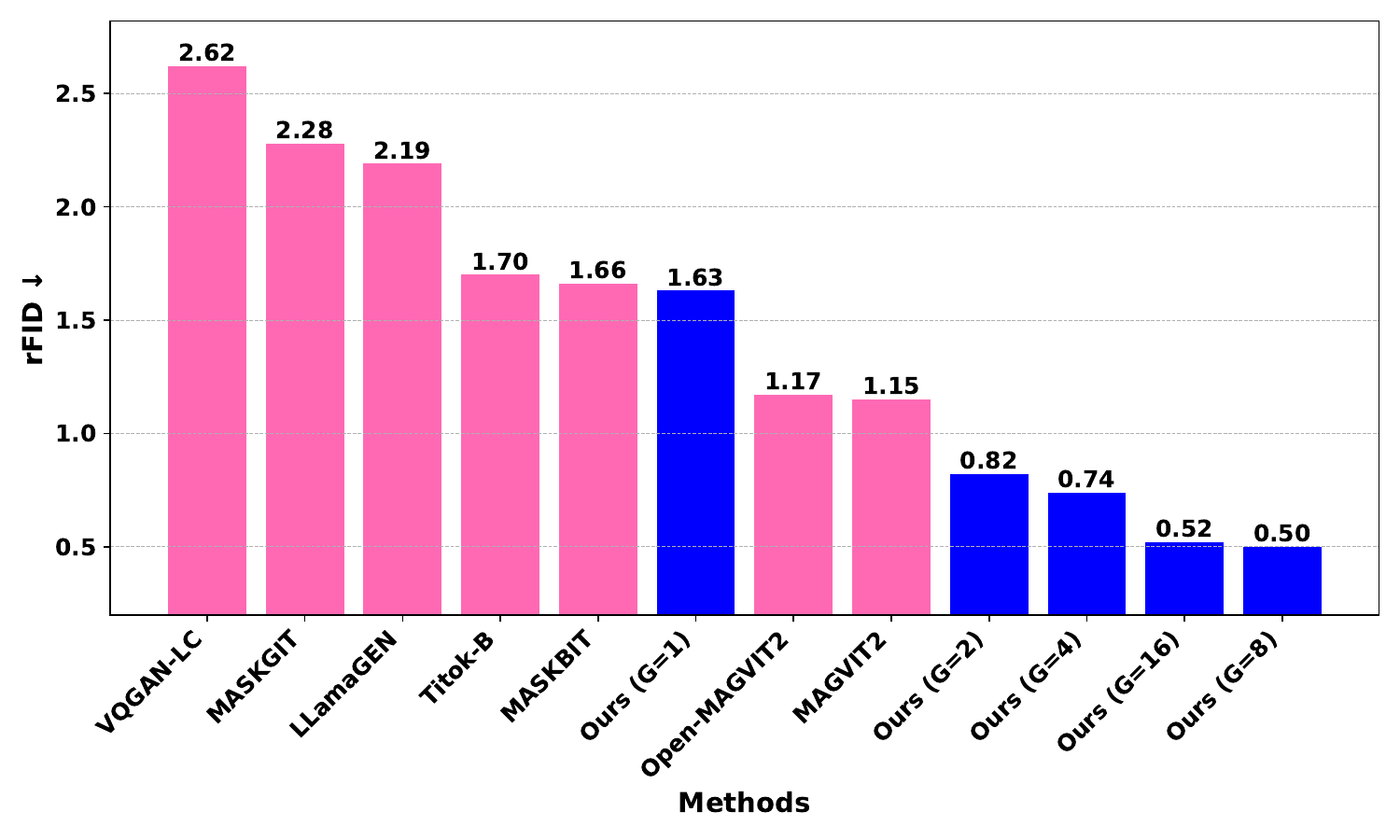}
    \subcaption{Reconstruction performance of \textbf{GSQ} with a latent dimension of \textbf{16} at \textbf{16$\times$} spatial compression, compared to the state-of-the-art. %
    }
    \end{minipage}\\
    \begin{minipage}[t]{0.6\textwidth}
        \centering
        \includegraphics[width=\textwidth]{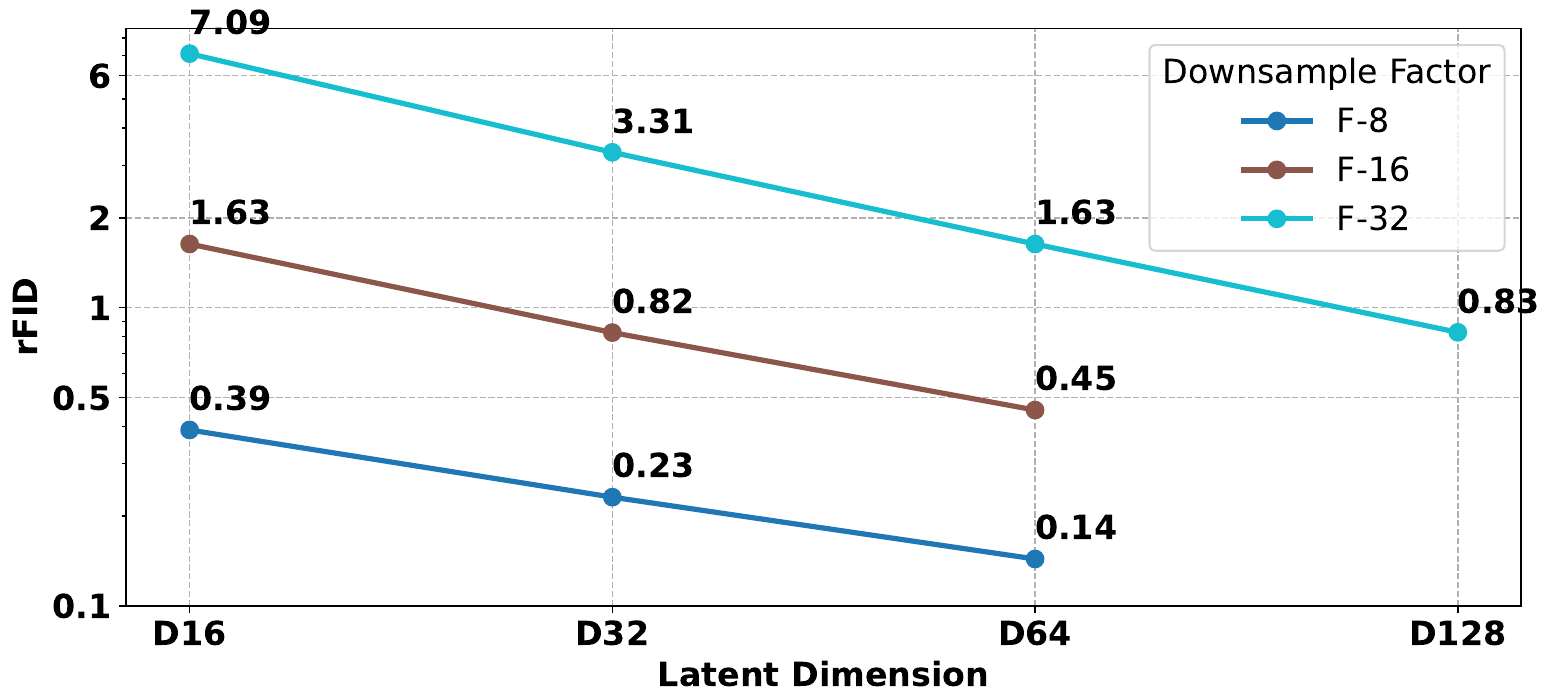}
        \subcaption{Scaling behaviour of the latent dimension v.s. spatial compression factor in \textbf{GSQ}; $d=16$ is fixed while groups $G$ increase to expand latent space.}
        \label{fig:exp_scaling_factor}
    \end{minipage}    

    \caption{
     The top figure shows \textbf{GSQ-GAN}’s reconstruction performance compared to state-of-the-art methods, demonstrating superior results even without latent decomposition. Training with larger $G$, which is more composed of groups, can further optimize the use of latent space, enhancing reconstruction quality. The bottom figure illustrates \textbf{GSQ-GAN}’s efficient scaling behaviour, where expanded latent capacity effectively manages increased spatial compression, thus achieving higher fidelity reconstructions on highly spatial compressed latent. Notably, \textbf{GSQ-GAN} achieves these results with only 20 training epochs on ImageNet at $256^2$ resolution, while methods, such as \cite{openmagvit2, magvit2}, require over 270 epochs.
    }
    \label{fig:gvq_vs_sota}

\end{figure}

Recent advancements in generative models for images and videos have seen substantial success, with approaches like autoregressive models \cite{llamagen, kondratyuk2023videopoet, wang2024emu3}, masked language models \cite{magvit2, yu2023magvit, chang2022maskgit, weber2024maskbit}, and diffusion-based methods (including score-matching and flow-matching) \cite{rombach2022high, yang2024cogvideox,hu2024zigma, gao2024lumin-t2x} and surpass GAN-based \cite{kang2023scaling, sauer2023stylegan-t} models. A common factor in many of these models is the reliance on latent discrete representations of images, especially within language-model-based, where continuous feature maps are quantized into discrete tokens. This quantization has become critical for high-fidelity generation, as tokenized images facilitate model efficiency and enhance generative quality, avoiding the need to work on high-resolution images directly. Recent studies \cite{magvit2, wang2024emu3} confirm
image tokenizer directly translates to generative quality, and the effectiveness of generative models is closely tied to the performance of image tokenizers.

The fundamental challenge in training image tokenizers is balancing compression efficiency with reconstruction accuracy. While recent methods show progress, several critical issues remain unresolved: (1) Many current tokenizers still depend on outdated GAN-based hyperparameters, often resulting in suboptimal, even negative, performance due to inconsistencies between generation and reconstruction objectives. (2) Benchmarking efforts frequently rely on legacy VQ-GAN implementations with outdated configurations, leading to biased comparisons and limited assessment accuracy. (3) Although various quantization models have been introduced, comprehensive analyses of their relative performance and scalability are limited, hindering the development of efficient, streamlined training methodologies for image tokenizers. Additionally, some methods, such as FSQ \cite{mentzer2023fsq} and LFQ \cite{magvit2},  rigidly bind latent dimension and codebook size, making independent scaling of either latent dimension or codebook size infeasible. To address these challenges, we propose the following contributions:
\begin{enumerate}
\item Grouped Spherical QuAnTization (GSQ): We introduce a novel approach featuring spherical codebook initialization and lookup regularization. With optimised configurations, GSQ outperforms state-of-the-art image tokenizers, achieving high performance with fewer training steps and without the need for auxiliary losses or GAN regularization.
\item Efficient Latent Space Utilization: GSQ achieves superior reconstruction performance with compact latent dimensions and large codebook sizes. Scaling studies reveal that latent space is often \textbf{underutilized} in lower spatial compression scenarios, underscoring the need for efficient latent space usage, which GSQ can address.
\item Scalability with Latent Dimensions: GSQ scales effectively with increasing latent dimensions by decomposing and grouping latents. Our spatial scaling studies indicate that latent space \textbf{saturation} occurs at larger spatial reduction scenarios. GSQ enables greater spatial reductions and leverages an expanded latent space to maximize the quantizer's capacity.
\end{enumerate}

These insights lay a foundation for more efficient and scalable training protocols in image tokenizers, advancing the potential of downstream tasks such as generative models for high-fidelity image generation tasks. We also demonstrate that our training approach can easily train up to $32\times$ spatial downsampling image tokenizer.

\section{Related Work}
\label{sec:literatures}

The Variational AutoEncoder \cite{kingma2013vae} is the foundational approach for image tokenization, initially developed to compress images into a continuous latent space, while later one more work focuses on refining continuous representations \cite{higgins2017beta, vahdat2020nvae, kim2019bayes, luhman2022optimizing, bhalodia2020dpvaes, egorov2021boovae, su2018f, qin2024epanechnikov}. Despite their strengths, however, these image encodings, often constrained by strong KL regularization, are rarely applied as image tokenizers within generative models. Instead, the VAE with vector quantization (VQ-VAE) \cite{vqvae, razavi2019generating} have become the preferred choice due to their effective use of a codebook for latent distribution regularization. Alternative variance is Residual Vector Quantizer (RVQ) \cite{zeghidour2021soundstream} that can achieve image compression and discrete quantization simultaneously.

Building on the success of VQ-VAE, the VQ-GAN model \cite{esser2021taming} further advanced image tokenizer training by incorporating a perceptual loss \cite{lpips} and an adversarial loss, enhancing the quality of generated images. Subsequent research has extended VQ-GAN through (1) architectural improvements, such as transformer-based structures \cite{vqgan-vit} and Layer Normalization \cite{chang2022maskgit}; (2) novel vector quantizers like Finite Scalar Quantization \cite{mentzer2023fsq}, Lookup-Free Quantizer \cite{magvit2} and so on \cite{bsq-vit,zheng2022high, zhu2024wavelet, sadat2024litevae, adiban2023s, yu2024spae, cao2023efficient, you2022locally, lee2022autoregressive, adiban2022hierarchical, kumar2024high, zheng2022movq, kumar2024high, li2024imagefolder, openmagvit2, VAR, fifty2024restructuringvectorquantizationrotation}; and (3) refined loss functions with perceptual enhancements, for example, using ResNet-based perceptual loss \cite{weber2024maskbit, yu2023magvit} and incorporating StyleGAN discriminators \cite{vqgan-vit, magvit2}. Our work primarily focuses on this stream of compression-oriented image tokenizer training, examining scaling behaviours and their influence on reconstruction quality.

An alternative line of research in image tokenization focuses on embedding semantic visual representations in the latent space, rather than maximising compression rates. This approach typically leverages pre-trained visual foundation models, such as DINO \cite{oquab2023dinov2}, CLIP \cite{radford2021clip}, and MAE \cite{he2022mae}, by transferring their learned representations into the latent space of image tokenizers or quantizing their latent representations. Early studies \cite{peng2022beitv2maskedimage, hu2023gaia1generativeworldmodel, Park_2023_ICCV_seit} demonstrated the feasibility of this strategy, though these models traditionally underperform in reconstruction quality compared to compression-driven tokenizers. Recent advancements have narrowed this gap by optimizing codebook initialization, refining network architectures, and employing advanced knowledge distillation methods, resulting in models that achieve competitive reconstruction fidelity while preserving strong semantic representation capabilities \cite{yu2024imageworth32tokens, zhu2024scalingcodebooksizevqgan, zhu2024stabilizelatentspaceimage, li2024imagefolder}.

\section{Methodology}
\label{sec:methdology}
\subsection{Preliminary: VQ Image Tokenizer }
The image tokenizer consists of an encoder $\operatorname{Enc}$ and a decoder $\operatorname{Dec}$. The encoder compresses the high-resolution input image $\mathbf{I} \in \mathbb{R}^{H \times W \times 3}$ into continuous latent maps:
\begin{equation}
    \mathbf{Z} = \operatorname{Enc}(\mathbf{I}) =\{ z_i\in\mathbb{R}^D\}_{i=1}^{h\times w}.
\end{equation} 
\noindent and the decoder reconstructs the image from the latent representation, $\hat{\mathbf{I}} = \operatorname{Dec}(\mathbf{Z})$. The down-sampling factor $f=\frac{H}{h}=\frac{W}{w}$ denotes spatial reduction, and the compression ratio is given by $\mathcal{R} = \frac{D}{3f^2}$, 
where $H$, $W$ is the height and width of input image $\mathbf{I}$ and $h$, $w$, $D$ is the height, width and dimension of latent. 

With a vector quantizer, the latent space is discretised by mapping $\mathbf{Z}$ to indices in the codebook $\mathbf{C} = \{ c_i\in \mathbb{R}^{D} \}_{i=1}^V $, where $V$ is the vocabulary size. Each latent vector $z_i$ from $\mathbf{Z}$ is quantized to the nearest codebook entry using a look-up operation, often based on Euclidean distance: 
\begin{equation} 
\operatorname{VQ}(z_i) = \operatorname{lookup}(z_i, \mathbf{C}) = \arg\min_{j}   ||z_i -c_j||^2.
\end{equation}

\subsection{Simple Scaling with GSQ}
Pursuing higher spatial reduction $f$ requires increasing the latent dimensionality $D$ to maintain $\mathcal{R}$, thus preserving reconstruction fidelity. However, increasing $D$ introduces high-dimensionality challenges, making distance computations less effective and limiting achievable compression ratios. One of the solutions is using a product quantizer  \cite{vahdat2020nvae, zheng2022high, zheng2022movq, jegou2010productquant}, hence we decompose each latent vector $z_i$ into $G$ groups: 
\begin{equation} 
\operatorname{GSQ}(z_i) = \{ \operatorname{lookup}^*(z_i^{(g)}, \mathbf{C}^{(g)}) \}_{g=1}^{G} ,
\end{equation}
Here, each $z_i^{(g)}$ represents a sub-group of $z_i$ with $d$ channels, where $G \times d = D$ enables efficient compression without compromising reconstruction fidelity. To improve stability and performance, we propose to initialize codebook entries from a spherical uniform distribution and same as \cite{vqgan-vit, bsq-vit}, apply $\ell_2$ normalization during lookup:
\begin{align} c^{(g)}_j &\sim \ell_2(\mathcal{N}(0, 1)), \\ 
\operatorname{lookup}^* (z_i, \mathbf{C}) &= \arg\min_{j} ||\ell_2(z_i) - \ell_2(c_j)||^2 .
\end{align}
We employ a shared codebook among all groups and omit $\ell_2$ when $G/D \in \{1, 2\}$, in which case GSQ reduces to LFQ \cite{magvit2}, and the spherical space significantly collapsed, which requires additional entropy loss during training \cite{magvit2, bsq-vit}. Further discussion is provided in Appendix \ref{sec:appendix_all_quantizers}.

\section{Experiments}
\label{sec:EXPERIMENT}

\subsection{Optimized Training for GSQ-VAE}
We first investigate the efficacy of our proposed improvements to GSQ on VAE-based tokenizers, including impacts of training configurations, auxiliary losses, model architecture, and hyperparameter settings. We set $G=1$ for all modes, they were trained on $128^2$ resolution ImageNet \cite{deng2009imagenet} with a down-sampling factor $f=8$, vocabulary size $V=8{,}192$, latent dimensionality $D=8$, with batch size 256, and learning rate of $1e^{-4}$  for 100k steps (20 epochs). Specific hyperparameters are reported in Appendix \ref{sec:appendix_exp_vae_setup}. All tokenizers adopted an exponential moving average with a decay rate of 0.999. We utilized the LPIPS perceptual loss \cite{lpips} as proposed in \cite{esser2021taming} with a weight of 1.0 in training. 
\begin{table}[hbt!]
\centering
\begin{tabular}{rc|ccccccc}
\bottomrule
 \textbf{Codebook Init} & \textbf{Norm}   & \textbf{rFID} $\downarrow$ & \textbf{IS} $\uparrow$  & \textbf{LPIPS} $\downarrow$  &  \textbf{PSNR} $\uparrow$  &  \textbf{SSIM} $\uparrow$   &  \textbf{Usage} $\uparrow$  & \textbf{PPL} $\uparrow$ \\ \hline
 $\mathcal{U}(-1 / V, 1 / V)$  &   &11.37 & 84 &0.12 &22.3 & 0.64 & 3.38\% &  237 \\ 
 $\mathcal{U}(-1 / V, 1 / V)$ & $\ell_2$ & 5.343 & 113 & 0.10 & 23.7 & 0.71   & 100\% &8077 \\ 
 $\ell_2(\mathcal{N}(0, 1))$  &    & 5.343 & 113 & 0.12 & 23.9 & 0.72   & 100\% &7408 \\ 
 $\ell_2(\mathcal{N}(0, 1))$  & $\ell_1$ & 8.312 & 94 & 0.12 & 22.1 &0.66  & 33.9\% & 566 \\ 
\rowcolor{gray!20}
  $\ell_2(\mathcal{N}(0, 1))$ &$\ell_2$ & 5.375 &113 &0.11 & 23.59 &0.71  & 100\%    &8062 \\ \hline
\bottomrule
\end{tabular}
\caption{Ablation of spherical codebook initialization and lookup normalization for \textbf{GSQ-VAE-F8} models, trained on ImageNet with $128^2$ resolution for 20 epochs. PPL is the perplexity. }
\label{tab:exp_vae_baseline}
\end{table}

\subsubsection{Effectiveness of Spherical Quantization}
\label{sec:res_1.1}
\textbf{Baseline and codebook initialization.} \cref{tab:exp_vae_baseline} demonstrates that our spherical uniform distribution codebook initialization significantly improved codebook usage to nearly 100\% during training. Using $\ell_2$ normalization, mentioned with previous studies \cite{vqgan-vit, bsq-vit}, is crucial for stabilizing codebook usage (especially in larger codebooks) and ensuring all codes are usually equal. As illustrated in \cref{fig:exp_codebook_init_codebook_usage}, our approach maintained approximately 100\% codebook utilization throughout training, which enabled the reduction of the rFID from 11.37 to 5.375, and with $\ell_2$ the perplexity of codebook usage is close to the vocabulary size.
\begin{figure}[hbt!]
    \centering
    \includegraphics[width=0.6\textwidth]{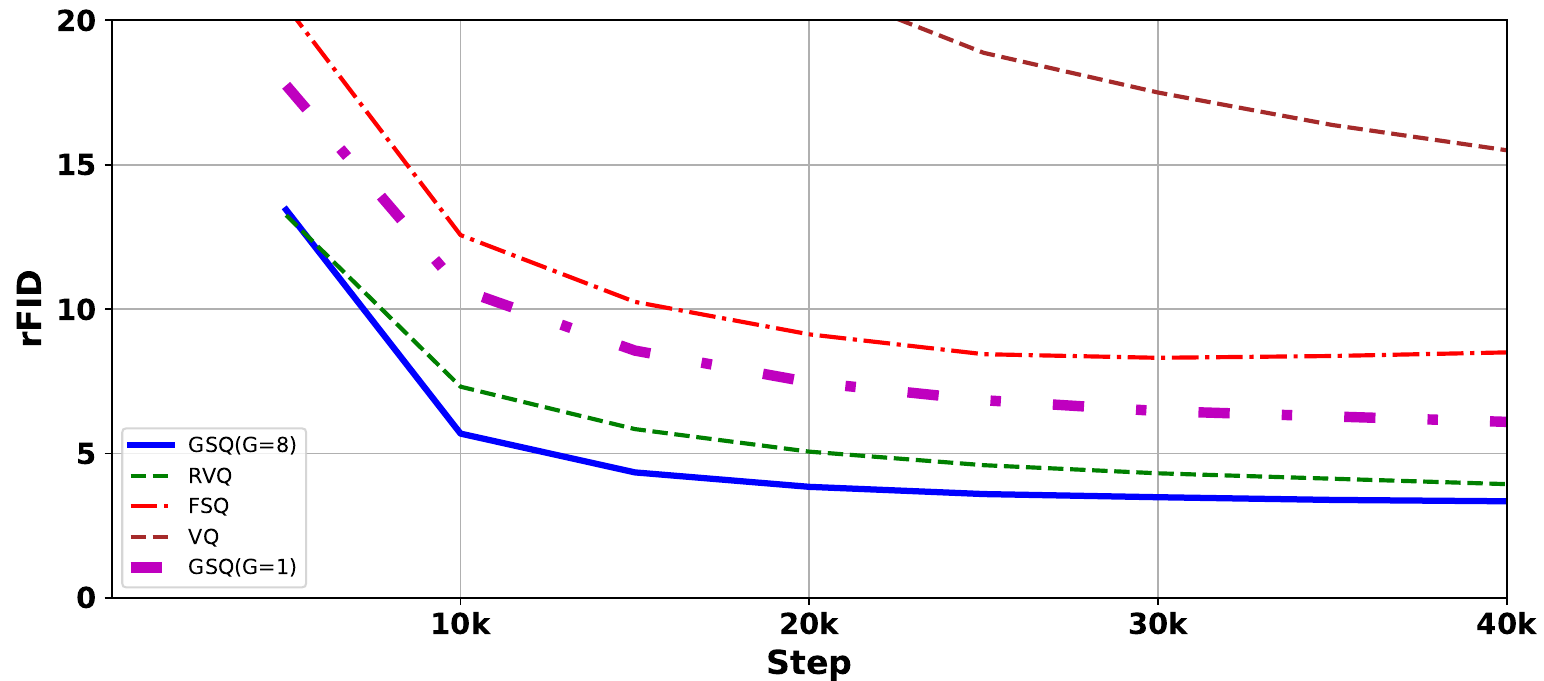}
    \caption{Comparisons of quantizers for \textbf{VAE-F8} training. VQ is initialized with uniform distribution; all models have the same backbone, latent dimension, and vocabulary size.  }
    \label{fig:exp_vae_quant_ablation}
\end{figure}

\paragraph{Quantizer Comparisons.} 
Taking the proposed spherical codebook initialization method and $\ell_2$ normalized lookup, GSQ (similar to VQ, when $G$ is 1) can outperform FSQ \cite{mentzer2023fsq}, and by scaling $G$ to 8, GSQ can beat RVQ \cite{zeghidour2021soundstream}, as we reported in \cref{fig:exp_vae_quant_ablation}, all model here has same latent dimension eight and vocabulary size 8,192.

\paragraph{Codebook auxiliary loss.} We investigated the effectiveness of codebook auxiliary losses, e.g. entropy loss \cite{magvit2, openmagvit2} and TCR loss \cite{zhang2023tcr}. \cref{tab:exp_vae_entropy} reveals that these losses negatively impacted the tokenizer performance and impeded codebook usage. Entropy loss only provided a marginal improvement with a minimal weight (0.01). Given their limited utility and computational cost on large vocab size during training, we opted not to use them. Also, the later results show that our method maintained 100\% codebook usage for vocabulary sizes up to 512k without these losses.

\begin{table}[hbt!]
\centering
\begin{tabular}{cc|ccccccc}
\bottomrule
\textbf{Entropy Loss} & \textbf{TCR Loss}   & \textbf{rFID} $\downarrow$ & \textbf{IS} $\uparrow$  & \textbf{LPIPS} $\downarrow$  &  \textbf{PSNR} $\uparrow$  &  \textbf{SSIM} $\uparrow$   &  \textbf{Usage} $\uparrow$  & \textbf{PPL} $\uparrow$   \\ \hline
0.01        &  & 5.281 & 114 & 0.12 & 23.9 & 0.72 & 99.8\% & 7397 \\ 
0.1         &  & 5.687 & 112 & 0.12 &23.7 &0.71 & 73.5\%  &5399 \\ 
0.5         &  & 7.906 & 97  &0.11  &22.8 &0.67 & 8.83\% &620 \\ 
& 0.01      & 9.937    & 82 &0.15 &22.5 & 0.65 & 81.1\% &830\\ 
\rowcolor{gray!20}
\ding{55} & \ding{55}   & 5.375 & 113 &0.11 & 23.59 &0.71 & 100\% &8062   \\ \hline

\bottomrule
\end{tabular}
\caption{Ablation of codebook auxiliary loss for \textbf{GSQ-VAE-F8}. Our methods enable the codebook usage to always be full; there is no need to use this auxiliary loss for training.}
\label{tab:exp_vae_entropy}
\end{table}

\subsubsection{Ablation of Network Backbone}
\label{sec:res_1.2}

We explored variations in baseline architectures, including the effect of Adaptive Group Normalization (as known as AdaLN) \cite{adain} and Depth2Scale \cite{magvit2}. As detailed in \cref{tab:exp_vae_network_arch}, surprisingly, these modules degraded the reconstruction's perceptual quality, increasing the rFID but decreasing the pixel-wise error. 
We use Adaptive Group Normalization as the default and further invested Depth2Scale in GAN's training in \cref{sec:exp_gan_depth2scale}.

\begin{table}[hbt!]
\centering
\begin{tabular}{cc|ccccccc}
\bottomrule
\textbf{AGN} & \textbf{Depth2Scale} & \textbf{rFID} $\downarrow$ & \textbf{IS} $\uparrow$  & \textbf{LPIPS} $\downarrow$  &  \textbf{PSNR} $\uparrow$  &  \textbf{SSIM} $\uparrow$   &  \textbf{Usage} $\uparrow$  & \textbf{PPL} $\uparrow$    \\ \hline
\rowcolor{gray!20}
\ding{55}   &\ding{55}   & 5.375 & 113 &0.11 & 23.59 &0.71 & 100\% &8062   \\ 
\rowcolor{gray!20}
\checkmark  &            & 5.406 &113 &0.10 & 23.85 &0.71 & 100\% &7457 \\ 
            & \checkmark & 5.562 &113 &0.11 &23.93 &0.72  & 100\% &7410 \\ 
\checkmark  & \checkmark & 5.531 &112 &0.11 &23.94 &0.72  & 100\% &7452 \\ \hline
\bottomrule
\end{tabular}
\caption{Ablation of using Adaptive Group Norm (AGN) and Depth2Scale for \textbf{GSQ-VAE-F8}. }
\label{tab:exp_vae_network_arch}
\end{table}

\begin{table}[hbt!]
\centering
\begin{tabular}{lcc|cccccccc}
\bottomrule
\textbf{Type} & $\mathbf{\lambda_p}$ & $\mathbf{\lambda_{rec}} $ & \textbf{rFID} $\downarrow$ & \textbf{IS} $\uparrow$  & \textbf{LPIPS} $\downarrow$  &  \textbf{PSNR} $\uparrow$  &  \textbf{SSIM} $\uparrow$   &  \textbf{Usage} $\uparrow$  & \textbf{PPL} $\uparrow$  \\ \hline
\multirow{5}{*}{LPIPS} 
     & 0.1  &  1.0 & 7.062 & 98 &0.12 & 25.26 &0.75 &100\% & 7013\\ 
      & 0.1 & 5.0 & 12.18  &73 &0.14 & 25.68 &0.75 &87\% & 5673\\ 
      &\cellcolor{gray!20}1.0  &\cellcolor{gray!20}1.0 & \cellcolor{gray!20}5.406 &\cellcolor{gray!20}113 & \cellcolor{gray!20}0.10 &\cellcolor{gray!20}23.85 &\cellcolor{gray!20}0.71  &\cellcolor{gray!20}100\% &\cellcolor{gray!20}7457        \\ 
      & 1.0  &  5.0 & 6.156 &105 &0,11 &24.93 &0.74  &100\%  &7192    \\ 
      & 10 &  1.0 & 6.093 &115 &0.11 &22.41  &0.68 &99\%  &7417    \\ 
\hline

\multirow{3}{*}{Dino} 
         & 0.1  &  1.0 & 7.312 &90 & 0.15 & 24.91 & 0.72 &100\% &6457 \\ 
         & 0.1  &  5.0 & 4.250&112 &0.12 &23.12 & 0.65 & 100\% &7004 \\ 
         & 0.7  &  4.0 & 4.343&110 &0.13 &23.66 &0.67 &100\% &6887\\ 
\hline
\multirow{3}{*}{ResNet} 
 & 0.1  &  1.0 & 31.37 &53 & 0.19 &21.70 & 0.57 &37\% &2657 \\ 
 & 0.1  &  5.0 & 9.625 &84 &0.15 &23.91  &0.68 & 73\% & 5001\\ 
 & 0.7  &  4.0 & 204 & 1.60 & 0.56 & 20.16 &0.41 &77\% &5028\\ 
\hline 
\multirow{3}{*}{VGG-16} 
 & 0.1  &  1.0 & 4.468 &112 &0.14 &22.64 &0.63 & 100\% &6926 \\
 & 0.1  &  5.0 & 5.031 &111 &0.14 &21.97 &0.61 &100\% &6986 \\ 
 & 0.7  &  4.0 & 4.906 &103 &0.15 &24.17 &0.69 &100\% &6759 \\ 
\hline
\bottomrule
\end{tabular}
\caption{Ablation of perceptual loss and weights for \textbf{VAE-F8} training. $\lambda_p$ and $\lambda_{rec} $ are weights of perceptual and reconstruction loss.}
\label{tab:exp_vae_ploss}
\end{table}

\subsubsection{Ablation of Perceptual Loss Selection}
\label{sec:res_1.3}
We explored various perceptual loss configurations, including LPIPS \cite{lpips} and logit-based perceptual loss with different backbone architectures: ResNet \cite{he2015deepresiduallearningimage}, VGG \cite{simonyan2015deepconvolutionalnetworkslargescale}, and Dino \cite{oquab2023dinov2}. As presented in \cref{tab:exp_vae_ploss}, our findings indicate that ResNet-based logit loss is ineffective as a perceptual loss, which contradicts earlier findings \cite{weber2024maskbit}. In contrast, Dino and VGG-based logit losses yielded lower rFID scores, demonstrating their potential. However, we opted for LPIPS due to its ability to effectively balance rFID and pixel-wise error. We anticipate that further optimisation through detailed hyperparameter tuning could enhance the performance of stronger perceptual losses.

\subsubsection{Hyper-parameters optimization for GSQ-VAE}
\label{sec:res_1.4}

\paragraph{Optimizers.} The choice of hyper-parameters specifically $\beta$ in Adam, significantly affects training dynamics. We evaluated combinations of $\beta$ values, ranging from 0 to 0.9, and reported results in \cref{tab:exp_vae_betas}. Our experiments reveal higher  $\beta$ always brings better reconstruction performance %
by promoting stable training. We also assessed weight decay values of $5e^{-2}$ and $1e^{-4}$, and results show that when higher $\beta$ is used, weight decay with $5e^{-2}$ performing best overall. Therefore, we use $\beta=[0.9, 0.99]$ with a weight decay of 0.05 for optimal training stability.

\begin{table}[hbt!]
\centering
\begin{tabular}{lc|cccccccc}
\bottomrule
$\beta$ & \textbf{Weight Decay} & \textbf{rFID} $\downarrow$ & \textbf{IS} $\uparrow$  & \textbf{LPIPS} $\downarrow$  &  \textbf{PSNR} $\uparrow$  &  \textbf{SSIM} $\uparrow$   &  \textbf{Usage} $\uparrow$  & \textbf{PPL} $\uparrow$    \\ \hline
\multirow{2}{*}{(0, 0.99)}    &$5e^{-2}$   & 5.562 & 113 &0.11 & 23.9 &0.72 &100\%&7410 \\ 
                              &$1e^{-4}$   & 5.812 & 107 &0.11 &23.9 & 0.71 &100\% &7393 \\ \hline
\multirow{2}{*}{(0.5, 0.99)}  &$5e^{-2}$   & 5.750 & 111 &0.10 & 23.85 & 0.71 & 100\% & 7492 \\
                              &$1e^{-4}$  &  5.375 & 109 &0.09 &23.85 &0.71 & 100\% &7421 \\ \hline
\multirow{2}{*}{(0.9, 0.95)}  &$5e^{-2}$  &  5.406 &113 &0.10 & 23.85 &0.71 & 100\% &7457 \\
                              &$1e^{-4}$  & 5.562 & 113 & 0.10 & 23.85 & 0.71 & 100\% &7407  \\ \hline
\multirow{2}{*}{(0.9, 0.99)}  &\cellcolor{gray!20} $5e^{-2}$   &\cellcolor{gray!20}5.343 &\cellcolor{gray!20}113 &\cellcolor{gray!20}0.10 &\cellcolor{gray!20}23.89 &\cellcolor{gray!20}0.71 &\cellcolor{gray!20}100\% &\cellcolor{gray!20}7462  \\
                              &$1e^{-4}$ & 5.562 & 112 &0.10 &23.86 &0.71 &100\% &7404 \\ \hline
\multirow{2}{*}{(0.9, 0.999)} &$5e^{-2}$   & 5.406 & 112 &0.10 & 23.87 &0.71 &100\% &7472  \\
                              &$1e^{-4}$ & 5.468 & 111 &0.10 & 23.88 &0.71 &100\%&7411  \\
\hline
\bottomrule
\end{tabular}
\caption{Optimizer's $\beta$ and weight decay ablations for \textbf{GSQ-VAE-F8} training. The codebook usage is 100\% for all models.}
\label{tab:exp_vae_betas}
\end{table}

\begin{table}[hbt!]
\centering
\begin{tabular}{lcc|ccccccc}
\bottomrule
\textbf{Warm-up} & \textbf{Decay} & \textbf{Final L.R.}  & \textbf{rFID} & \textbf{IS}   & \textbf{LPIPS}  &  \textbf{PSNR}  &  \textbf{SSIM} &  \textbf{Usage}  & \textbf{PPL}  \\ 
 &  &  &  $\downarrow$ & $\uparrow$  & $\downarrow$  &  $\uparrow$  & $\uparrow$   &  $\uparrow$  & $\uparrow$ \\ \hline
\rowcolor{gray!20}
0     & \ding{55}  &  1$e^{-4}$  & 5.343 &113 &0.10 &23.89 &0.71 &100\% &7462  \\
5k    & \ding{55}  &  1$e^{-4}$  & 5.406 & 114 &0.10 &23.78 &0.72 & 100\% &7429 \\ %
5k & 75k & $1e^{-5}$ & 5.750 & 110 &0.10 &23.67&0.71 &100\%&7344   \\ %
5k & 95k & $1e^{-5}$ & 5.781 & 109 &0.09 &23.76 &0.71 &100\% &7355    \\ %
5k & 95k & 0 & 5.625 & 111 &0.10 & 23.73 & 0.71 &100\% &7343 \\ %
5k & 10\% at 75k & $1e^{-5}$ & 5.468 & 112 & 0.10 &23.83 &0.71 &100\% &7389 \\ %
\hline
\bottomrule
\end{tabular}
\caption{Learning rate scheduler ablations for \textbf{GSQ-VAE-F8} training, the maximal learning rate is 1$e^{-4}$. The codebook usage is 100\% for all models.}
\label{tab:exp_vae_lr_sched}
\end{table}

\paragraph{Learning rate scheduler.} Recent studies used diverse learning rate schedulers for training tokenizers. We compared fixed learning rate training against the other five schedulers, each with a 5k steps warm-up period and varied decay strategies, as plotted in \cref{fig:exp_vae_lr_sched} in Appendix \ref{sec:appendix_exp_vae_setup}. The results are reported in \cref{tab:exp_vae_lr_sched}, showing that substantial learning rate decay negatively impacted model performance, and there are no advantages from warm-up training. Therefore, we opted for a constant learning rate throughout training to maintain the GAN training and simplicity of hyper-parameter optimization.

\subsection{Optimized Training for GSQ-GAN}
Next, we incorporated a discriminator and adversarial loss to ablate training configurations for GSQ-GAN training on ImageNet \cite{deng2009imagenet} at $128^2$ resolution for up to 80k steps; the VAE and discriminator have a learning rate of $1e^{-4}$. Detailed hyperparameters are reported in Appendix \ref{sec:appendix_exp_gan_setup}.

\begin{table}[hbt!]
\centering
\begin{tabular}{lcc|cccccc}
\bottomrule
\textbf{Discriminator} & \textbf{Adv.} & \textbf{Discr.}  & \textbf{rFID} & \textbf{IS}     &  \textbf{PSNR} &  \textbf{SSIM} &  \textbf{Usage}& \textbf{PPL}    \\ 
 & \textbf{loss} & \textbf{loss}  & $\downarrow$ &$\uparrow$& $\uparrow$  &$\uparrow$  &  $\uparrow$  &  $\uparrow$ \\ \hline
\ding{55}     & \ding{55}  &  \ding{55} & 5.343 & 113 &23.89 &0.71   & 100\% &7462\\ \hline

\multirow{6}{*}{\begin{tabular}[c]{@{}l@{}}NLD \\ \cite{patchgan}\end{tabular}}
   & Hinge & Vanilla                  &  45.2 &25&20.6 &0.58  &96.4\% & 6976 \\ 
   & Hinge & Hinge                    & 24.0 &49  &21.4 &0.62  &98.5\% &7424 \\ 
   & Hinge & Non-Sat.           & 68.5  &14 &19.3 &0.51 &58.2\% &4069 \\ 
   &\cellcolor{gray!20}Non-Sat. &\cellcolor{gray!20}Vanilla &\cellcolor{gray!20}9.562 &\cellcolor{gray!20}86 &\cellcolor{gray!20}22.08 &\cellcolor{gray!20}0.66   &\cellcolor{gray!20}100\% &\cellcolor{gray!20}7558\\ 
   & Non-Sat. & Hinge           & 11.3 &80 & 22.0 &0.66   &100\% & 7516\\ 
   & Non-Sat. & Non-Sat.  & 23.7 &50 &21 &0.62  &99.0\% &7451 \\ \hline

\multirow{3}{*}{\begin{tabular}[c]{@{}l@{}}SGD \\(1k) \\ \cite{stylegan}\end{tabular}}
   & Hinge & Hinge       & 18.1  & 63 &21.65 & 0.64 & 100\%  &6104  \\ 
   & Non-Sat. & Vanilla  & 19.1  & 62 &21.57 &0.64 & 100\%  &6061     \\ 
   & Non-Sat. & Hinge    & 27.1  & 46 &21.42 &64.96 & 100\%  &5514 \\ \hline

\multirow{3}{*}{\begin{tabular}[c]{@{}l@{}}DD \\ \cite{sauer2023stylegan-t}\end{tabular}}
  & Hinge & Hinge             &1.976 &116 &21.78 &0.64 & 100\% & 7546 \\
  & Non-Sat. & Vanilla       &1.906 &117 &22.01 &0.65 & 100\% & 7533 \\
  &\cellcolor{gray!20}Non-Sat. &\cellcolor{gray!20}Hinge     &\cellcolor{gray!20}1.867 &\cellcolor{gray!20}117 &\cellcolor{gray!20}22.12 &\cellcolor{gray!20}0.66 & \cellcolor{gray!20}100\% &\cellcolor{gray!20}7525 \\ \hline
\multicolumn{3}{c|}{OpenMagViT2 w/ 1.75M steps} &  1.180 & \multicolumn{3}{c}{\cite{openmagvit2}}  \\
\hline
\bottomrule
\end{tabular}
\caption{\textbf{GSQ-GAN-F8} model trained on 128$^2$ ImageNet, 80k training step. The SGD-GAN model is evaluated at the 1k training step due to the failure of $\mathit{NaN}$ loss in training.}
\label{tab:exp_gan_loss}
\end{table}

\subsubsection{Ablations of Discriminator and Combinations of Adversarial Loss}
\label{sec:res_2.1}
We evaluated three types of discriminator: N-Layer Discriminator (\textit{NLD}) \cite{patchgan}, StyleGAN Discriminator (\textit{SGD}) \cite{stylegan}, and Dino Discriminator (\textit{DD}) \cite{sauer2023stylegan-t}. We also compared three adversarial loss types: vanilla non-saturating (\textit{V}), hinge (\textit{H}), and improved non-saturating (\textit{N}), resulting in six combinations of adversarial-discriminator loss setups. 

Choosing an improper GAN loss led to negative performance for N-Layer and Dino Discriminators. As shown in  \cref{tab:exp_gan_loss}. All GAN models trained with Dino Discriminators consistently outperformed GAN with the N-Layer one. The best losses for N-Layer Discriminator are with \textit{NV} losses, achieving an rFID of 9.562, and \textit{NH} for Dino Discriminator, which reached 1.867 rFID. 
Additionally, we ablate the data augmentation \cite{sauer2023stylegan-t} in Dino Discriminator, as shown in \cref{tab:exp_gan_dd_aug}, using a combination of colour augmentation, translation, and cutout led to improved reconstruction performance.

\begin{table}[hbt!]
\centering
\begin{tabular}{l|cccc}
\bottomrule
\textbf{Discr. Data Aug.} &\textbf{rFID-128}$^2$ $\downarrow$ &\textbf{rFID-256}$^2$  $\downarrow$  \\ \hline
\ding{55}          & 1.953 & 0.824 \\
Color+Trans        & 2.000 & 0.783  \\
\rowcolor{gray!20} 
Cutout+Color+Trans & 1.867  & 0.824   \\
Resize+Color+Trans & 2.000 & 0.832   \\ \hline
\bottomrule
\end{tabular}
\caption{Ablation on data augmentation in Dino-Discriminator.}
\label{tab:exp_gan_dd_aug}
\end{table}

\begin{table}[hbt!]
\centering
\begin{tabular}{lccc|cccc}
\bottomrule
\textbf{Discr.} & \textbf{Loss} & $\beta$ & $\lambda_{adv}$ &\textbf{rFID} $\downarrow$ & \textbf{IS} $\uparrow$   &  \textbf{PSNR} $\uparrow$  &  \textbf{SSIM} $\uparrow$  \\ \hline
NLD & NH & (0, 0.99)    & 0.1 & 6.687 & 96.5 &22.35 &0.67  \\
NLD & NH & (0.5, 0.9)  & 0.1 & 11.31 & 80.0 &22.01 &0.66 \\
NLD & NH & (0.5, 0.9)  & 0.5 & 106 & 8.68  &15.40 & 0.29 \\
NLD & NH & (0.9, 0.95)  & 0.1 & 3.578 & 114 &22.74 &0.69 \\
NLD & NH & (0.9, 0.99)  & 0.1 & 3.515 & 114 &22.85 &0.69  \\ 
NLD & NH & (0.9, 0.99)   & 0.5 & 3.718 & 114 &22.83 &0.69\\ 
\hline
NLD & NV &  (0.5, 0.9) & 0.1 & 9.562 & 86 & 22.08 & 0.66  \\
\rowcolor{gray!20}
NLD & NV &  (0.9, 0.99) & 0.1 & 3.390 & 102 &22.88 &0.69 \\
NLD & NV & (0.9, 0.99)  & 0.5 & 3.515 & 114 &22.86 &0.69 \\

\hline
DD & NH & (0.5, 0.9)   & 0.1 & 1.867 & 117 &22.12 &0.66\\
\rowcolor{gray!20} 
DD & NH & (0.9, 0.99)   & 0.1 & 1.859 & 118 &22.12 &0.66 \\
DD & NH & (0.9, 0.99) & 0.5 & 2.453 & 106 & 20.66 &0.59 \\
DD & NV & (0.5, 0.9)   & 0.1 & 1.906 & 117 &22.01 &0.65 \\
DD & NV & (0.9, 0.99)   & 0.1 & 1.820 & 117 & 22.02 & 0.65 \\
DD & NV & (0.9, 0.99) & 0.5 & 2.671 & 102 &20.28 &0.57  \\ \hline

\bottomrule
\end{tabular}
\caption{Ablation of Adam's $\beta$ and adversarial loss weights for \textbf{GSQ-GAN-F8} training. $\lambda_{adv}$ is the weight of adversarial loss.}
\label{tab:exp_gan_optimizers}
\end{table}

\subsubsection{Hyper-parameters Optimization for GSQ-GAN}
\label{sec:res_2.2}
\textbf{Discriminator optimizer and adversarial loss weights.} 
We performed ablation studies on optimizer hyper-parameters ($\beta$) for N-Layer and Dino Discriminator. The results, presented in \cref{tab:exp_gan_optimizers}, indicate that higher $\beta$ values ($\beta = [0.9, 0.99]$) led to more stable training dynamics for both discriminator types. We used this configuration for the remainder of the experiments. Additionally, varying the weight of adversarial loss did not show significant benefits, leading us to set the adversarial loss weight to 0.1.

\begin{table}[hbt!]
\centering
\begin{tabular}{cc|cccccccc}
\bottomrule
\textbf{Batch size} & \textbf{Learning rate}  & \textbf{rFID} $\downarrow$ & \textbf{IS} $\uparrow$  & \textbf{LPIPS} $\downarrow$  &  \textbf{PSNR} $\uparrow$  &  \textbf{SSIM} $\uparrow$   &  \textbf{Usage} $\uparrow$  & \textbf{PPL} $\uparrow$    \\ \hline
256 & 1$e^{-4}$  & 1.859 & 118 & 0.08 & 22.12 & 0.66 & 100\% &7528  \\
\rowcolor{gray!20} 
256 & 2$e^{-4}$  & 1.796 & 119 &0.07 &22.28 &0.66 &100\% &7525  \\
256 & 3$e^{-4}$  & 1.890 & 118 &0.07 &22.36 &0.67 &100\% &7544  \\
512 & 1$e^{-4}$  & 1.671 & 120 &0.08 &22.08 &0.66 &100\% &7494   \\
\rowcolor{gray!20} 
512 & 2$e^{-4}$  & 1.578 & 122 &0.07 &22.25 &0.66 &100\% &7538   \\
\rowcolor{gray!20}
768 & 2$e^{-4}$  & 1.593 & 121 &0.07 &22.32 &0.67 &100\% &7513   \\
768 & 3$e^{-4}$  & 1.648 & 122 &0.07 &22.31 &0.67 &100\% &7520 \\ 
\hline
\bottomrule
\end{tabular}
\caption{Batch size and learning rate ablations of \textbf{GSQ-GAN-F8} training, with \textbf{DD-NH} discriminator and loss combination.}
\label{tab:exp_gan_bs_LR}
\end{table}

\paragraph{Learning Rates and Batch Size.}
We investigated the batch size and learning rate configurations, comparing three different batch sizes and learning rates. The results, shown in \cref{tab:exp_gan_bs_LR}, indicate that larger batch sizes and increased learning rates improved stability and convergence speed and thus allowed us to speed up GAN training with larger batch sizes. %

\subsubsection{GAN Regularization Ablations}
\label{sec:res_2.3}
We explored several regularization techniques for stabilizing discriminator training: gradient penalty \cite{gulrajani2017improvedtrainingwassersteingans}, LeCAM regularisation \cite{yu2023magvit}, and autoencoder warm-up, as well as adaptive discriminator loss weights \cite{vqgan-vit}, weight decay, and gradient clipping. \cref{tab:exp_gan_regularzation} summaries our findings.

Using constant $\lambda_{adv}$ performed best, with no advantages observed from adaptive weighting \cite{esser2021taming}. The Gradient penalty added for N-Layer Discriminator was ineffective, and LeCAM only slightly improved results. Autoencoder warm-up (discriminator training starts after 20k steps) did not improve stability or performance; gradient clipping at 2.0 (by default) was more effective than at 1.0, and weight decay of $1e^{-4}$ improved the N-Layer Discriminator but slightly degraded the Dino Discriminator. 

Training StyleGAN Discriminator with regularization could not address $\mathit{NaN}$ issues. We also tested a combination of StyleGAN Discriminator and gradient penalty. But training with gradient penalty was also roughly four times slower, so we could not finish the training within 80k step training wall time, see more details of StyleGAN Discriminator in Appendix \ref{sec:appendix_exp_gan_setup}).

\begin{table}[hbt!]
\centering
\begin{tabular}{lcc|ccccccc}
\bottomrule
\textbf{Discr.} & \textbf{WD} & \textbf{AW}    &\textbf{rFID} $\downarrow$ & \textbf{IS} $\uparrow$  & \textbf{LPIPS} $\downarrow$  &  \textbf{PSNR} $\uparrow$  &  \textbf{SSIM} $\uparrow$& \textbf{PPL} $\uparrow$  \\ \hline
\rowcolor{gray!20} 
NLD-NV & $5e^{-2}$    &                      & 3.390 &114 &0.06 & 22.8 & 0.69  & 7594 \\
NLD-NV + GC 1.0 & $5e^{-2}$    &             & 3.453 &114 &0.06 &22.8 & 0.69  &7483\\
NLD-NV & ${1e^{-4}}$  &                      &\textcolor{red}{3.296} & 115 & 0.06 & 22.86 & 0.69  &7494 \\
NLD-NV & $5e^{-2}$    & \checkmark           & 4.437 &112 &0.07 &23.34 &0.70  &7476 \\
NLD-NV + GP & $5e^{-2}$    &                 & 5.750 &110 &0.09 &23.78 &0.71  &7447\\
NLD-NV + LeCAM & $5e^{-2}$     &             & 3.546 &113 &0.07 &22.89 &0.69  &7455 \\ \hline
\rowcolor{gray!20} 
DD-NH & $5e^{-2}$     &                      & 1.859 & 118 & 0.08 & 22.12 & 0.66  &7528 \\
DD-NH & ${1e^{-4}}$     &                    & 1.914 &118  &0.08  & 22.12 & 0.66 &7514 \\
DD-NH & $5e^{-2}$     & \checkmark           & 2.687 &117 &0.07&23.40 &0.70  &7464\\
DD-NH + AE-warmup & $5e^{-2}$         &      & 2.000 &116 &0.08 &22.22 & 0.66  &7484 \\

DD-NH + LeCAM & $5e^{-2}$    &               & 5.250 &111 & 0.08 &23.79 &0.71 & 7437\\ \hline

\rowcolor{gray!20} 
SGD-NH & $5e^{-2}$     &  \checkmark         & 3.593 &110 & 0.07 & 23.61 &0.70  &7470\\ 

\hline
\bottomrule
\end{tabular}
\caption{Ablation studies of GAN's regularization technologies for \textbf{GSQ-GAN-F8} training, WD is weight decay, AW is adversarial loss adaptive weight \cite{esser2021taming}, GC is gradient clip. All modes are trained with gradient clip 2.0 by default, GP is gradient penalty, and LeCAM's weight is 0.001 if enabled; when warmup is used, the discriminator starts to be updated after 20k iterations.   }
\label{tab:exp_gan_regularzation}
\end{table}

\begin{table}[hbt!]
\centering
\begin{tabular}{ccccc}
\bottomrule
\textbf{Data Aug}    & \textbf{D2S}               & \textbf{Attention}                 & \begin{tabular}[c]{@{}c@{}}\textbf{rFID}$\downarrow$\\ \textbf{128}\end{tabular} & \begin{tabular}[c]{@{}c@{}}\textbf{rFID}$\downarrow$\\ \textbf{256}\end{tabular} \\ \hline
           &             &             & 1.609                   & 0.675 \\
\checkmark &             &             & 1.578                   & 0.652 \\
           & \checkmark  &             & 1.570              & 0.660 \\
\rowcolor{gray!20}      
\checkmark & \checkmark  &             & 1.531             & 0.605\\
           &             & \checkmark  & \textcolor{red}{1.421}    & 0.605 \\
\checkmark &             & \checkmark  & 1.539            & 0.585 \\
           & \checkmark  & \checkmark  & 1.523         & 0.660 \\
\hline
\multicolumn{3}{c}{OpenMagViT2 \cite{openmagvit2} w/ 1.75M steps}    &1.180 & 0.34	 \\
\hline
\bottomrule
\end{tabular}
\caption{Ablation of discriminator data augmentation,  integration of attention and Depth2Scale for \textbf{GSQ-GAN-F8} training. D2S is the short for Depth2scale. }
\label{tab:exp_gan_attention}
\end{table}

\subsubsection{Analysis of Attention Integration}
\label{sec:res_2.4}
\label{sec:exp_gan_depth2scale}
We conducted ablation studies on the attention module and Depth2Scale layers. Recent works such as \cite{openmagvit2, magvit2} omit attention layers, but as seen in \cref{tab:exp_gan_attention}, incorporating attention into mid-blocks improved model performance. We also re-evaluated Depth2Scale, observing that it enhanced GAN’s performance under adversarial training. The model's performance across different resolutions is also reported in \cref{tab:exp_gan_attention}, showing the model's cross-resolution inference capabilities. We take Depth2Scale, as it generally benefits the GAN training; the model trained with Depth2Scale has rFID 1.53 with 80k training steps. Including attention modules can further boost reconstruction, though it may introduce instability during training.

\vspace{-1cm}
\subsection{Scaling Behaviors of GSQ-GAN }
This section investigates how variations influence reconstruction quality in latent dimensions and codebook vocabulary size. All models in this study were trained at a $256^2$ resolution with a batch size of 512 over 50k steps (20 epochs). Detailed hyper-parameters are provided in Appendix \ref{sec:appendix_scaling_latent}.

\begin{figure}[hbt!]
    \centering
    \includegraphics[width=.55\textwidth]{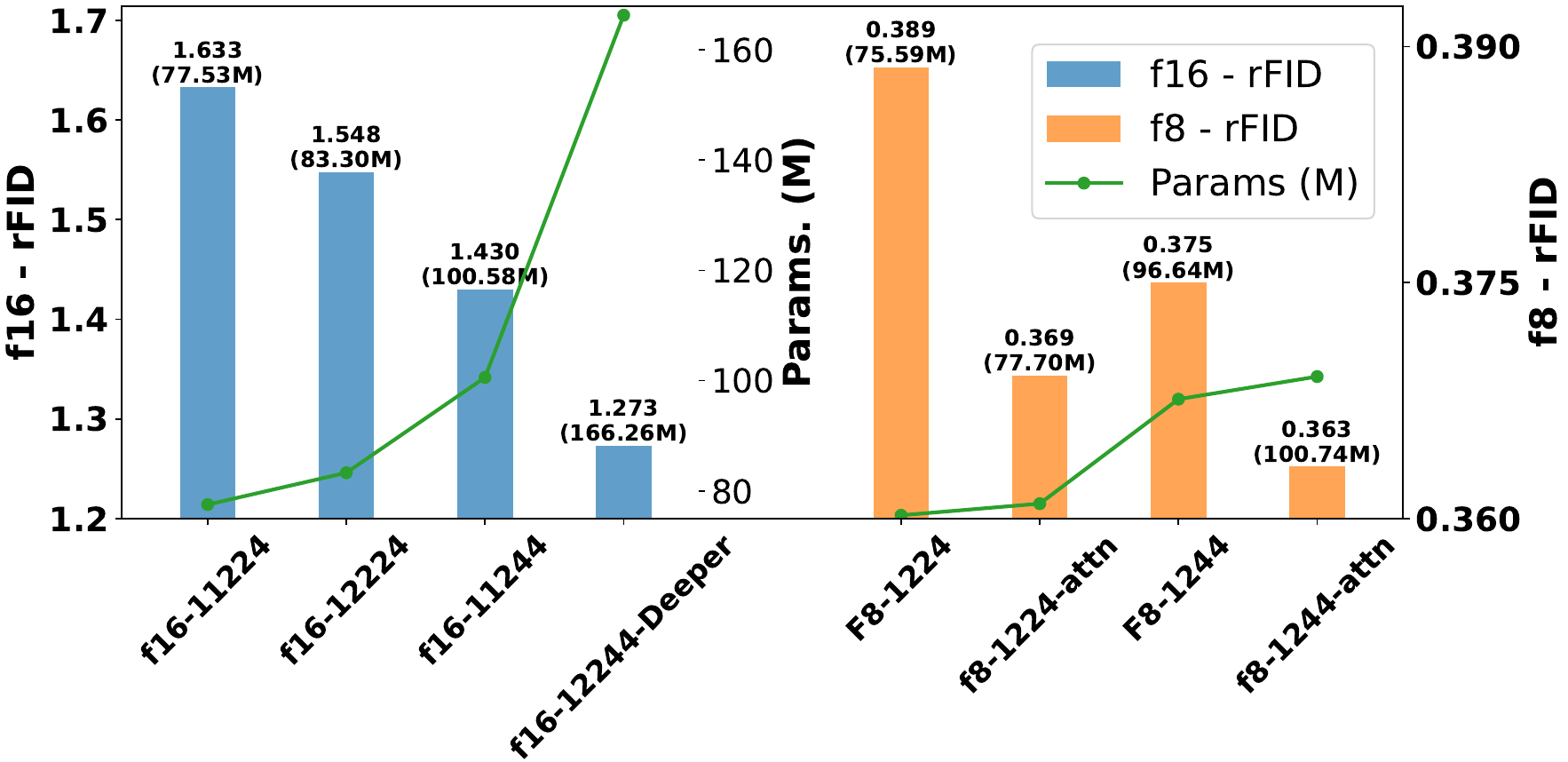}
    \caption{\textbf{GSQ-GAN} ablations on wider and deeper networks w/ and w/o attention blocks. Models are trained on $256^2$ resolution on ImageNet.}
    \label{fig:exp_scaling_network}
\end{figure}

\subsubsection{Network Capacity. }
\label{sec:res_3.1}
We examine the effects of network capacity on reconstruction fidelity, specifically looking at the width and depth. Width scaling was implemented by increasing the number of channels in convolution layers, while depth scaling involved adding additional convolution blocks\cite{magvit2}. The results, summarized in \cref{fig:exp_scaling_network}, demonstrate consistent improvements in reconstruction as network width and depth increase. Integrating attention modules within wider networks yielded further gains as used in \cite{esser2021taming}.

\begin{figure}[hbt!]
    \centering
    \begin{minipage}[t]{0.7\textwidth}
        \centering
        \includegraphics[width=\textwidth]{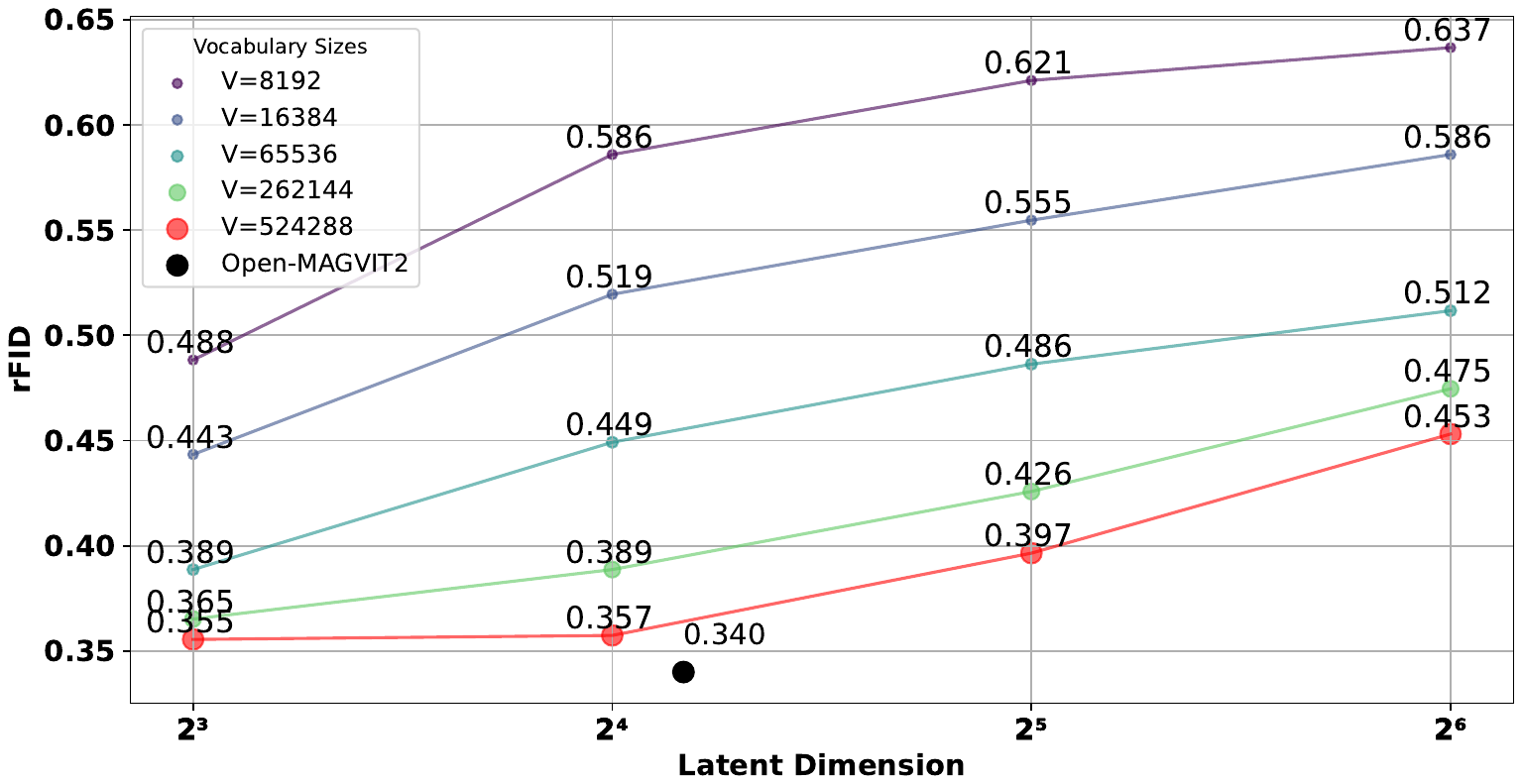}
        \subcaption{Scaling of latent dimension and vocabulary size for \textbf{GSQ} at \textbf{8$\times$} spatial compression.}
        \label{fig:exp_scaling_latent}
    \end{minipage}    \\
    \begin{minipage}[t]{0.7\textwidth}
    \centering
    \includegraphics[width=\textwidth]{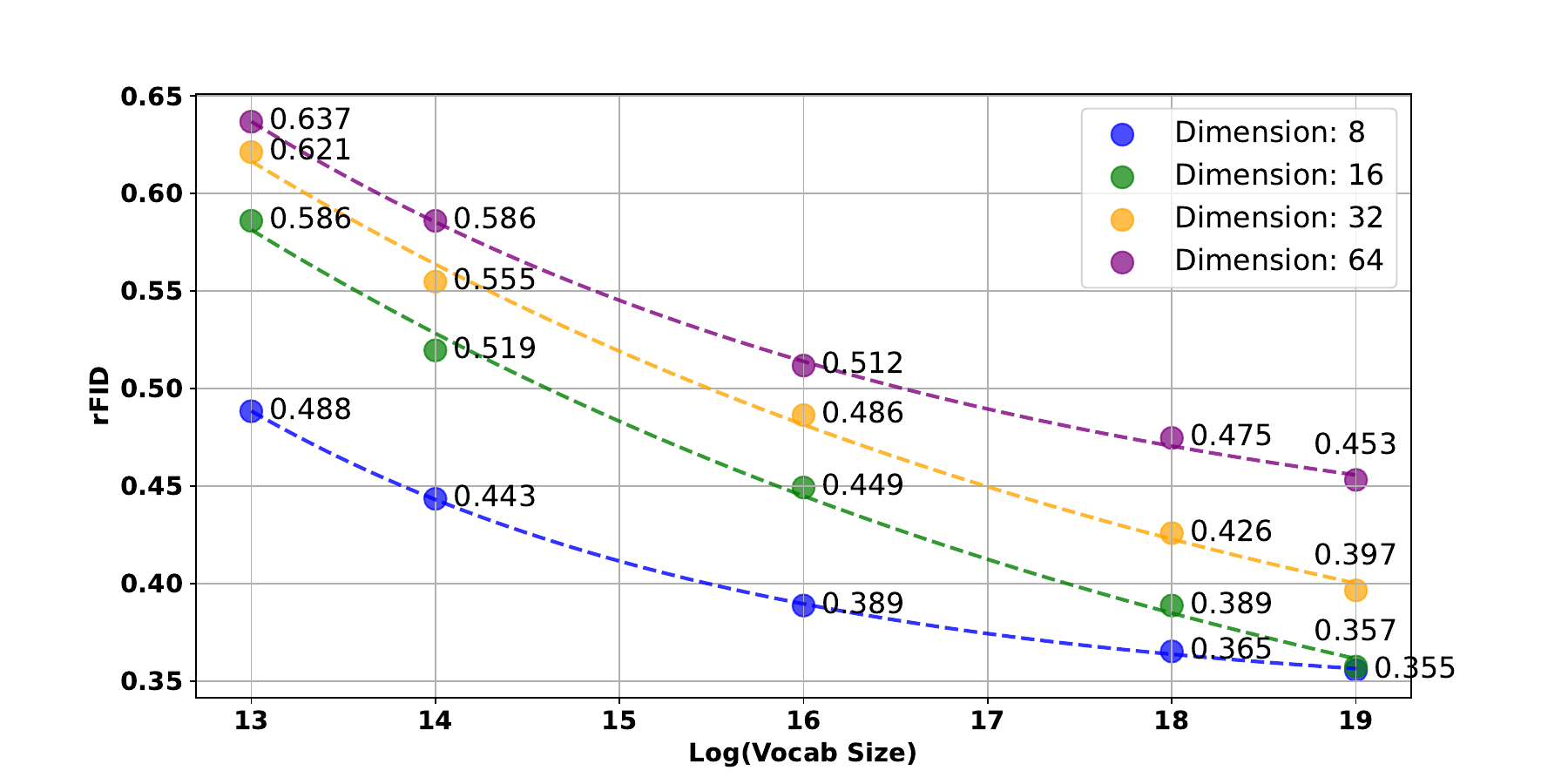}
    \subcaption{Same scaling behaviour as the top figure with vocabulary size in logarithmic scale.}
    \label{fig:exp_scaling_latent_logv_fid_fitiing}
    \end{minipage}
    \caption{The top figure illustrates the scaling of latent dimension and codebook size for \textbf{GSQ} at \textbf{8$\times$} spatial compression, where a smaller latent dimension improves reconstruction, suggesting the latent space is \textbf{not saturated} for \textbf{F8} downsampling. Optimising latent space size further enhances performance. The bottom figure shows the same trend with vocabulary size in logarithmic scale, indicating effective scaling as vocabulary size increases. All models are trained with \textbf{$G=1$} and no latent decomposition, making this equivalent to VQ-based methods. All models are trained on ImageNet at $256^2$ resolution.}
    \label{fig:the_exp_scaling_latent_and_vocab}
\end{figure}

\subsubsection{Scaling  of Latent Space and Vocabulary.}
\label{sec:res_3.2}
Next, we investigate the impact of scaling latent dimensionality and codebook vocabulary size. Models were trained with latent dimensions of $2^3$, $2^4$, $2^5$, and $2^6$, each paired with vocabulary sizes of 8k, 16k, 64k, 256k, and 512k. Results in \cref{fig:exp_scaling_latent} and \cref{fig:exp_scaling_latent_logv_fid_fitiing} indicate that larger vocabulary sizes, combined with lower latent dimensions, consistently yielded superior reconstruction performance. Remarkably, a model with a latent dimension of 8 and a vocabulary size 512k outperformed the state-of-the-art image tokenizers, achieving notable results within just 50k training steps (20 epochs).

These findings underscore the significance of a large codebook vocabulary in enhancing quantizer representational capacity. This trend aligns with theoretical expectations, as the representational capacity of GSQ-GAN is fundamentally bounded by $\log V$ as shown in \cref{fig:exp_scaling_latent_logv_fid_fitiing}, where $V$ is the vocabulary size. The pattern holds consistently across configurations and provides a point of contrast with prior studies with VQ (e.g., \cite{magvit2} \cite{vqgan-vit} \cite{llamagen}), as they did not employ optimized configurations for VQ-GAN training that the model training degradation has a bias on their scaling behaviours observation.

Our experiments reveal that lower-dimensional latent spaces result in improved reconstruction fidelity. As detailed in Appendix \ref{sec:appendix_all_quantizers}, low-dimensional latent spaces are advantageous for computing precise Euclidean distances used for codebook updates. This insight supports the success of decomposed vector quantization approaches, such as LFQ \cite{magvit2}, FSQ \cite{mentzer2023fsq}, and our own proposed GSQ.%

\begin{figure}[hbt!]
    \centering
    \begin{minipage}[t]{0.6\textwidth}
        \centering
        \includegraphics[width=\textwidth]{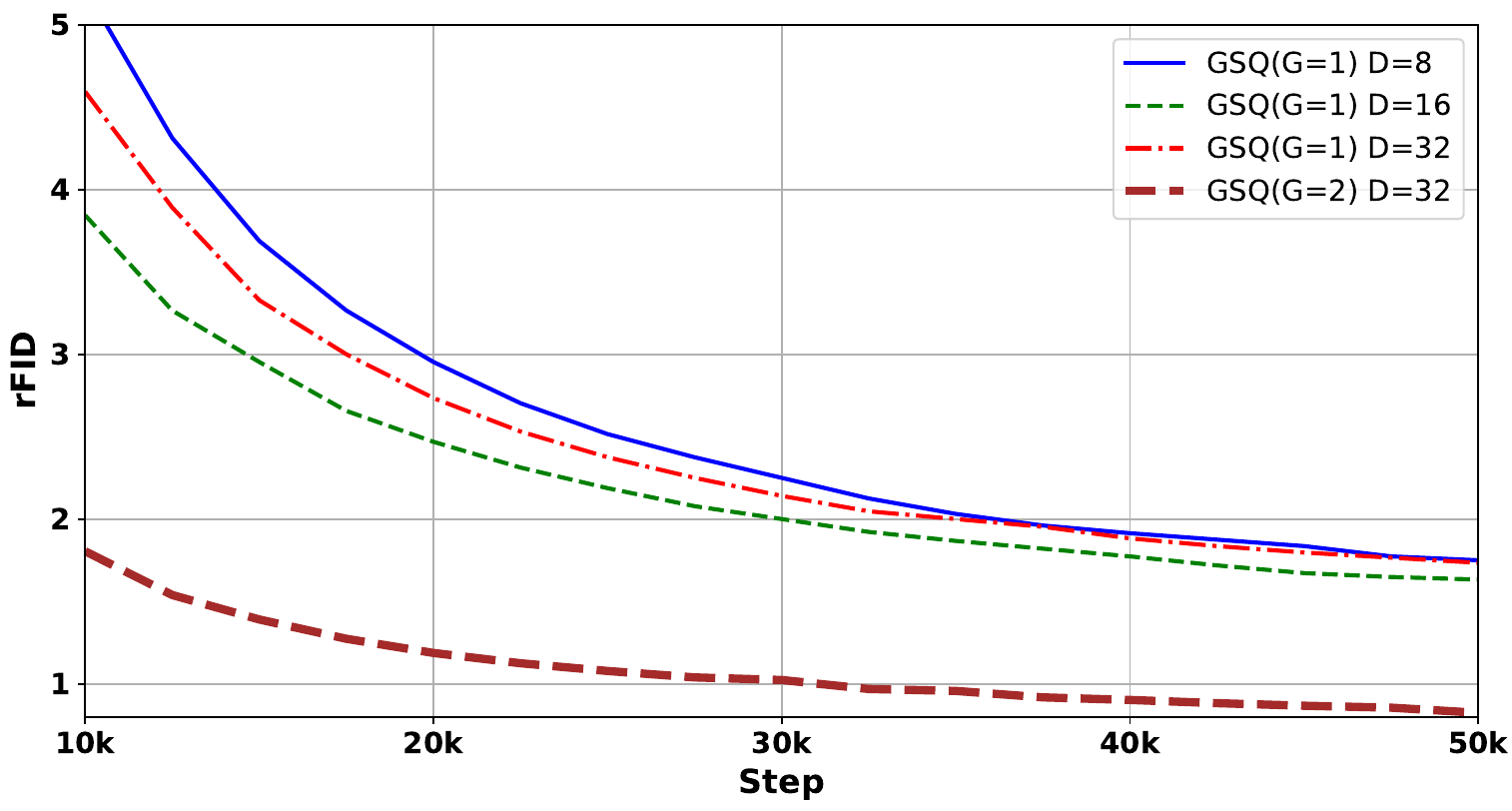}
    \end{minipage}
    \caption{Latent dimension scaling for \textbf{GSQ-GAN-F16} training, the latent space is \textbf{saturated} for \textbf{F16} spatial compression; we expect to enhance reconstruction performance by increasing the latent dimension to increase the latent capacity. Only \textbf{GSQ} with latent decomposition can scale to a higher latent dimension. }
    \label{fig:exp_vq_downsameple_sacing}
\end{figure}

Interestingly, one might intuitively expect a larger latent dimension to yield better performance because of the huge latent space. Our results suggest that high-dimensional spaces are often underutilized. This is important since effective compression at higher spatial down-sampling ratios requires larger latent dimensionality. However, normal VQ-like models cannot effectively scale latent dimensions against high spatial compression challenges. As illustrated in \cref{fig:exp_vq_downsameple_sacing}, increasing latent dimensionality enhances reconstruction quality when moving from F8 to F16. However, beyond a certain point (here is F16-D16), the model encounters the well-known limitations imposed by the curse of dimensionality. By contrast, when using the dimension decomposition in GSQ, even with $G=2$, the reconstruction performance gains fascinating improvement.  

\subsubsection{Latent Space and Downsample Factor, and Better Scaling with GSQ}
\label{sec:res_3.3}
To address the limitations regarding the difficulty of scaling attend dimension. We use GSQ to decompose large latent dimensions into low dimensions, thus maximizing reconstruction fidelity more effectively. As demonstrated in \cref{tab:exp_gvq_g_ablation}, by decomposing latent vectors into multiple groups, GSQ significantly enhances reconstruction performance without changing the overall latent dimensionality or vocabulary size. This result confirms GSQ’s ability to harness the representational power of high-dimensional latent spaces, leading to substantial gains in model fidelity. 

Notably, the model achieves near-lossless reconstruction with $D=64$ and $G=16$, approaching theoretical maximum performance. Although the compression ratio is very low and lacks practical value, it highlights GSQ’s remarkable scalability and representational power.

\begin{table}[hbt!]
\centering
\begin{tabular}{l|r|ccccccc}
\bottomrule
                     \textbf{Models}    & $G\times d$& \textbf{rFID} $\downarrow$ & \textbf{IS} $\uparrow$  & \textbf{LPIPS} $\downarrow$  &  \textbf{PSNR} $\uparrow$  &  \textbf{SSIM} $\uparrow$   &  \textbf{Usage} $\uparrow$  & \textbf{PPL} $\uparrow$ \\ \hline
\begin{tabular}[l]{@{}l@{}}\cite{openmagvit2} \\LFQ F16-D18 \\ $V=256 \text{k}$ \end{tabular}  & $18 \times 1$ & 1.17   \\ \hline

\multirow{4}{*}{\begin{tabular}[c]{@{}l@{}}GSQ F8-D64 \\ $V=8 \text{k}$\end{tabular}} 
                              & $1 \times 64$ & 0.63 &205 & 0.08 & 22.95 &0.67 & 99.87\% &8,055  \\
                              & $2 \times 32 $ & 0.32  &220 &0.05 & 25.42 & 0.76 & 100\% &8,157   \\
                              & $4 \times 16 $ & 0.18  &226 &0.03 &28.02 &0.08 &100\% &8,143      \\
                              & \cellcolor{blue!20}  $16 \times4 $ &\cellcolor{blue!20}0.03 & \cellcolor{blue!20}233 &\cellcolor{blue!20}0.004 &\cellcolor{blue!20}34.61 &\cellcolor{blue!20}0.91 &\cellcolor{blue!20}99.98\% &\cellcolor{blue!20}6,775      \\
\hline

\multirow{4}{*}{\begin{tabular}[c]{@{}l@{}}GSQ F16-D16 \\ $V=256 \text{k}$\end{tabular}} 
                         & $1 \times 16$ & 1.63  &179 & 0.13 &20.70 &0.56 &100\% &254,044 \\
                         & $2 \times 8 $ & 0.82  &199 & 0.09 &22.20 &0.63 &100\% &257,273   \\
                         & $4 \times 4 $ & 0.74  & 202 &0.08 &22.75 &0.63 &62.46\% &43,767       \\
                         & $8 \times 2 $ & 0.50  & 211 & 0.06 &23.62 &0.66 &46.83\% &22,181      \\
                         & $16 \times 1 $ & 0.52 & 210 & 0.06 &23.54 &0.66 &50.81\% &181        \\
                         & $16 \times 1^{*} $ & 0.51 & 210 & 0.06 &23.52 &0.66 &52.64\% &748        \\
\hline
\multirow{4}{*}{\begin{tabular}[c]{@{}l@{}}GSQ F32-D32 \\ $V=256 \text{k}$\end{tabular}}
                         & $1 \times 32$ & 6.84 &95 &0.24  &17.83 &0.40 &100\% & 245,715    \\ 
                         & $2 \times 16$ & 3.31 &139 &0.18 &19.01 & 0.47 & 100\% & 253,369    \\
                         & $4 \times 8 $ & 1.77 &173 &0.13 &20.60 &0.53 &100\% &253,199   \\
                         & $8 \times 4 $ & 1.67 &176 &0.12 &20.88 &0.54 &59\% &40,307    \\ 
                         & $16 \times 2 $ & 1.13 & 190 & 0.10 &21.73 &0.57 &46\% &30,302   \\ 
                         & $32 \times 1 $ & 1.21 & 187 &0.10 &21.64 &0.57 &54\% &247   \\ 
                         \hline
\bottomrule
\end{tabular}
\caption{Ablation studies of group decomposition with 8, 16 and 32 spatial downsample, vocabulary size is 8k, 256k and 256k respectively. \textbf{GSQ} outperforms LFQ with $3\times$ lower rFID. $G$ is the number of groups, and $d$ is a latent dimension in each group. $16 \times 1^{*}$ is trained with clip instead of $\ell_2$ normalization.}
\label{tab:exp_gvq_g_ablation}
\end{table}

\paragraph{Scaling Down-sample Factor.}  With GSQ optimizing latent space utilization, we further investigate the impact of varying down-sampling factors on reconstruction quality. We conducted experiments across different configurations of latent dimensions and down-sampling factors. As illustrated in \ref{fig:exp_scaling_factor}, models trained with a down-sampling factor of $f=8/16/32$ showed a consistent improvement in reconstruction as latent dimensions increased (with $d=16$ and group count $G$ adjusted accordingly). These results align with theoretical expectations and further validate the effectiveness of GSQ in fully utilizing the latent space.

\section{Conclusion}
\label{sec:conclusion}

We introduce a novel quantization method, Grouped Spherical Quantization(GSQ), incorporating spherical codebook initialization, lookup normalization, and latent decomposition. We systematically investigate training strategies and optimizations for the proposed GSQ-GAN, identifying key configurations that enhance reconstruction quality with significantly fewer training iterations. We highlight critical scaling behaviours related to the model, latent space, and codebook vocabulary size, emphasizing the role of compact latent spaces in achieving high-fidelity reconstruction. Our results demonstrate that GSQ efficiently scales in high-dimensional latent spaces,  leverating latent decomposition and spherical normalization for improved compression and reconstruction. %

\section{Acknowledgment}

In alphabetical order, we thank Erik, Ismail, Jan, Lijun, and Oleg for their insightful input and feedback on this manuscript. This work was supported by the German Federal Ministry for Economic Affairs and Climate Action under the project “NXT GEN AI METHODS: Generative Methods for Perception, Prediction, and Planning,” the bidt project KLIMA-MEMES, Bayer AG, and the German Research Foundation (DFG) project 421703927. We also appreciate the Gauss Centre for Supercomputing e.V. for granting access to computing resources on the JUWELS and JURECA supercomputers at the Jülich Supercomputing Centre (JSC). The German AI Service Centre WestAI provided additional computational resources.

\bibliography{tmlr}
\bibliographystyle{tmlr}

\appendix
\clearpage
\onecolumn %

\tableofcontents
\newpage
\onecolumn %
\title{
    \centering
    \Large
    \textbf{\thetitle}\\
    \vspace{1.0em}
}
\maketitle

\appendix

\section{Performance of State-of-the-Art Image Tokenizers}
We list additional comparisons of reconstruction performance among various state-of-the-art image tokenizers, including the model trained in this study. The evaluation is conducted on ImageNet at a resolution of $256\times256$.
\begin{table*}[hbt!]
\centering
\begin{tabular}{llllll}
\bottomrule
                      & $f$               & Latent-size & $D$                 & $V$  & rFID \\ \bottomrule
Ours-GSQ               & 8                & $32\times 32$       & 8 ($d=8$,$G=1$)   & 8k            & 0.48 \\
Ours-GSQ               & 8                & $32\times 32$       & 8 ($d=8$,$G=1$)   & 256k          & 0.36 \\
Ours-GSQ               & 8                & $32\times 32$       & 16($d=16$,$G=1$)  & 256k          & 0.51 \\ 
Ours-GSQ               & 8                & $32\times 32$       & 64($d=4$,$G=16$)  & 8k            & 0.03 \\
\hline
VQ-GAN \cite{esser2021taming}               & 8                 & $32\times 32$       &                     & 8k            & 1.49 \\
VQGAN-LC  \cite{zhu2024scalingcodebooksizevqgan}            & 8                 & 1024        & 8                   & 100,000       & 1.29 \\ 
VIT-VQGAN\_SL  \cite{vqgan-vit}       & 8                 & $32\times 32$       & 32                  & 8k            & 1.28 \\
OmniTokenizer \cite{wang2024OmniTokenizer}         & 8                 & $32\times 32$       & 8                    & 8k           & 1.11 \\
OmniTokenizer \cite{wang2024OmniTokenizer}        & 8                 & $32\times 32$       & 8                    & $\infty$     & 0.69 \\
LlamaGen   \cite{llamagen}           & 8                 & $32\times 32$       & 8                   & 16k           & 0.59 \\
BSQ    \cite{bsq-vit}               & 8                 & $32\times 32$       & 36                  & $2^{36}$      & 0.41 \\
Open-MAGVIT2 \cite{openmagvit2}         & 8                 & $32\times 32$       & 18                  & 256k          & 0.34 \\
\bottomrule
\bottomrule
Ours-GSQ w/ attention & 16              & $16\times 16$       & 8($d=1$,$G=8$)       & 512k          & 0.95 \\ 
Ours-GSQ              & 16              & $16\times 16$       & 16($d=16$,$G=1$)     & 256k          & 1.42 \\ 
Ours-GSQ              & 16              & $16\times 16$       & 16($d=1$,$G=16$)     & 256k          & 0.52 \\ \hline 
VQGAN-LC  \cite{zhu2024scalingcodebooksizevqgan}   & 16              & 256         & 8                   & 100,000       & 2.62 \\
MASKGIT \cite{chang2022maskgit}              & 16              & $16\times 16$       & 256                 & 1k            & 2.28 \\
LlamaGen \cite{llamagen}   & 16              & $16\times 16$       & 8                   & 16k           & 2.19 \\
Titok-B \cite{yu2024imageworth32tokens}  & 16              & 128                 &                     & 4k            & 1.70 \\
MASKBIT \cite{weber2024maskbit}              & 16              & $16\times 16$       & 256                 & 1024          & 1.66 \\
ImageFolder  \cite{li2024imagefolder}   & 16              & 265                 &                     & 4k            & 1.57   \\
MAGVIT2 \cite{magvit2}& 16              & $16\times 16$       & 18                  & 256k          & 1.15 \\
Open-MAGVIT2\cite{openmagvit2}& 16              & $16\times 16$       & 18                  & 256k          & 1.17 \\ \hline
\bottomrule
\end{tabular}
\caption{Reconstruction performance comparison of the proposed model against other state-of-the-art methods on ImageNet ($256\times256$ resolution).}
\end{table*}

\section{Networks}
\begin{figure}[hbt!]
    \centering
    \begin{minipage}[t]{0.9\textwidth}
        \centering
        \includegraphics[width=\textwidth]{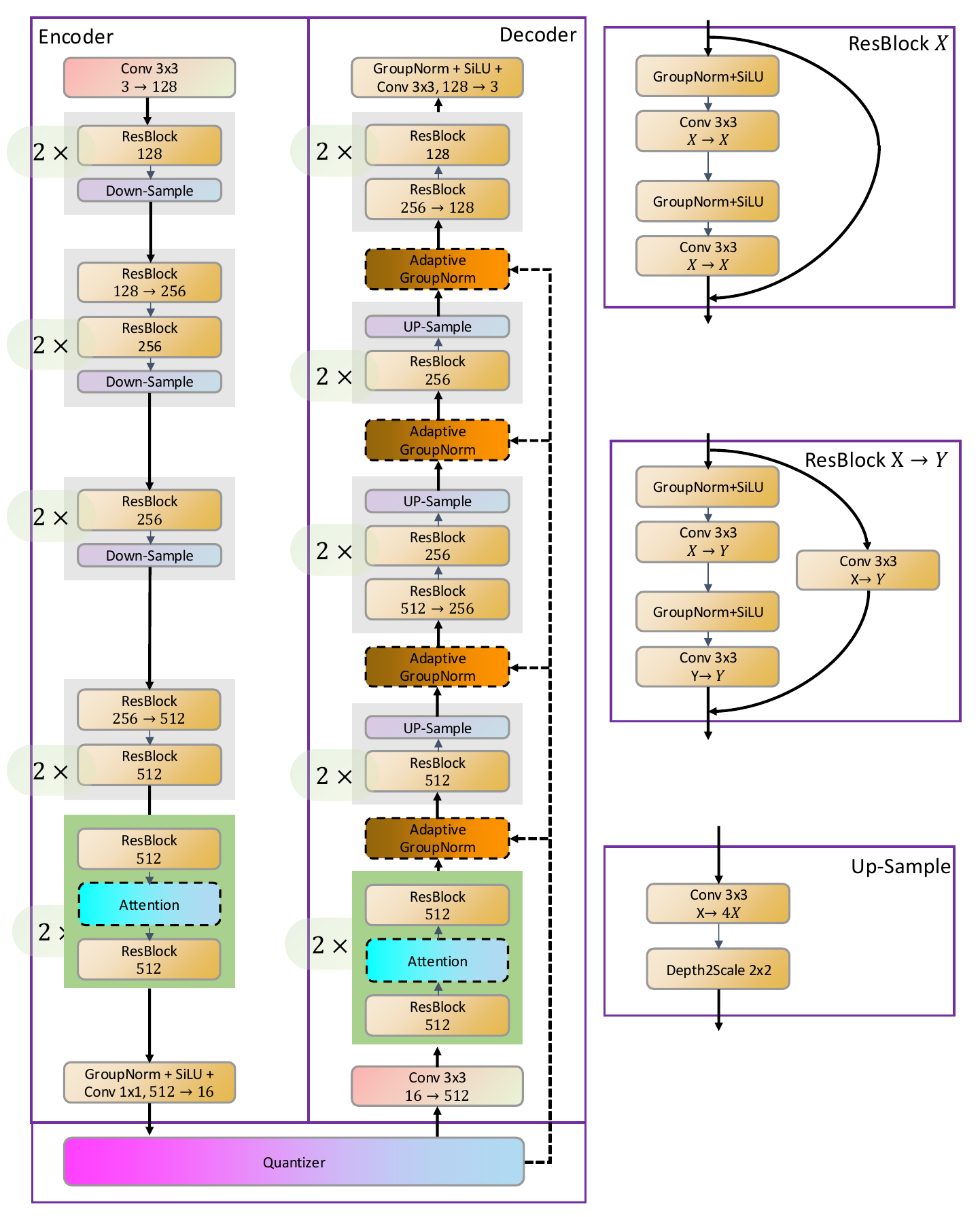}
    \end{minipage}
    \caption{Architecture of the GSQ tokenizer. The backbone follows the 2D convolutional version of MagVit2 \cite{magvit2}, with variations in the number of blocks.}
    \label{fig:networks}
\end{figure}

The network backbone is derived from VQ-GAN \cite{vqgan-vit}, and MagVit2 \cite{magvit2}. The encoder and decoder backbones are classified into two primary components: up/down-sampling resolution blocks (grey blocks in \cref{fig:networks}) and mid-blocks (green blocks in \cref{fig:networks}). We build down-sampling resolution blocks in the encoder with such rules:
1) For a spatial down-sampling factor of $f=2^N$, the encoder includes $N+1$ down-sampling blocks, each containing a ResBlock. The first $N$ blocks are followed by a down-sampling operation using stride convolutions. 2) the convolutional channels in each ResBlock and the number of ResBlocks within each down-sampling block are determined by the \textit{Channel Multipliers} and \textit{Encoder Layer Configurations}. 3) an additional ResBlock is introduced to match the channel dimensions if the channel multiplier doubles at a specific layer.
The decoder follows analogous principles, adding Adaptive GroupNorm layers before each up-sampling operation.

For mid-blocks each of the mid-blocks consists of a specified number of ResBlocks, with their channel dimensions determined by the output channels of the preceding layer. When mid-block attention mechanisms are used, attention is inserted between any two consecutive ResBlocks within the mid-blocks.

\newpage

\section{GSQ and Other Quantizers}
This section discusses the relationship between GSQ and other tokenizers. GSQ provides a unified framework for tokenizers, excluding the specific spherical codebook initialization proposed in this work. Other tokenizers can be derived by appropriate configurations, as outlined in \cref{tab:transfer_gsq_to_others}.
\label{sec:appendix_all_quantizers}
\begin{table}[hbt!]
\centering
\begin{tabular}{l|l|l|l|c|c|c|c|c|c}
\toprule
                 & $D$   & $d$    & $g$  & $V$      & Finite       & Codebook-Sharing         & $\ell_2$  & Fixed-Codebook   & Effective $V$            \\ \hline
VQ   & $D$   &  $D$   &  1   &  $V$     &  \ding{55} &  \ding{55}           & \ding{55}    & \ding{55}     &$V$         \\ \hline
VQGAN-ViT  & $D$   &  $D$   &  1   &  $V$     &  \ding{55} &   \ding{55}           &  \checkmark  & \ding{55}     &$V$     \\ \hline
LFQ   & $D$   &  1   & $D$ &  2       &  $\{-1, 1\}$  &    \ding{55}     &  \checkmark   &  \checkmark   &$2^D$      \\ \hline
FSQ   & $D$   &  1 &  $D$ &  $|\mathbf{C}^{(g)}|$     &  \checkmark  &  \ding{55} &  \checkmark &  \checkmark   &  $\prod_{g \in G} |\mathbf{C}^{(g)}|$           \\ \hline
BSQ    & $D$   &  2     & $\frac{D}{2}$ &    $V$       &    \ding{55}     &  \checkmark  &  \checkmark &  \checkmark  &  $V^\frac{D}{2}$ \\ \hline
GSQ       & $d\times g$   &  $d$  &  $g$   &  $V$     &  \ding{55} & $d > 2$  & \checkmark & \ding{55}  & $V^g$  \\ \bottomrule
\end{tabular}
\caption{The effective configurations of other tokenizers in GSQ's view.}
\label{tab:transfer_gsq_to_others}
\end{table}

\paragraph{VQ} VQ \cite{vqvae} and GSQ are identical when the latent space is not decomposed into groups ($G=1$) and without $\ell_2$ normalization.

\paragraph{BSQ} BSQ \cite{bsq-vit} represents the $d=2$ case of GSQ, where the number of groups is set as $G=\frac{D}{2}$. Codebooks are shared across groups, and BSQ's codebook is fixed.

\paragraph{FSQ} FSQ \cite{mentzer2023fsq} is a specific case of GSQ, where $G=D$, and each group has its own unshared, finite codebook. The term "finite" here refers to a small vocabulary size $V$. with each latent variable $z$ in the codebook $\mathbf{C}^{(g)}$ constrained as follows:
\begin{equation}
    Sigmoid(z) \in \{0, \frac{1}{V-1}, \frac{2}{V-1}, ..., 1\}
\end{equation}

In FSQ, typical values for $V^{(g)}$ are 5, 6, 7, or 8, representing a very small vocabulary size.

\paragraph{LFQ} LFQ \cite{mentzer2023fsq} can be interpreted from multiple perspectives. Within the GSQ framework, the simplest interpretation is to set $d=1$. For any 1-dimensional latent variable $z_i$, the $\ell_2$ normalization reduces to two possible outputs, $-1$ or $1$:
\begin{equation}
    \ell_2(z_i) = \frac{z_i}{||z_i||_2} =     \begin{cases}
      &  1, \text{if} \; z_i > 0 \\
      &  -1, \text{if} \; z_i < 0
    \end{cases}       
\end{equation}

Special cases, such as $z_i=0$, are handled by setting $\ell_2(z_i)=-1$ in alignment with \cite{mentzer2023fsq}. In this scenario, the 1-dimensional sphere degenerates into two discrete points, reducing the vocabulary size $V$ to 2. Prior studies \cite{magvit2, openmagvit2, bsq-vit} have shown the necessity of additional auxiliary objectives, such as entropy loss, to ensure effective codebook usage during training. However, in LFQ, codebook indices are not explicitly used; instead, the computational cost is transferred to entropy calculations. For large codebooks, even modern entropy computation kernels introduce significant memory and computational overhead.

There are two possible ways to address these challenges for $d=1$ with a shared codebook: Avoid applying $\ell_2$ normalization, thereby eliminating the vocabulary size degradation and the need for entropy loss and expensive entropy computations in large codebooks. Alternatively, we can enable $\ell_2$ normalization but use different codebooks among groups (very similar to LFQ). Both approaches generalize the 1-dimensional sphere into a 1-dimensional manifold, equivalent to the $d=2$ case of GSQ without $\ell_2$ normalization. We take the first solution in for $d=1$ case.

\subsection{Discussion of Euclidean Distance }
The squared Euclidean distance between an $n$-dimensional vector $z$ and a vector $\mathbf{C}$ in the codebook is given by: \begin{equation} || z - \mathbf{C} ||_2^2 = ||z||_2^2 + ||\mathbf{C}||_2^2 - 2(z \cdot \mathbf{C}), \end{equation} where $z \cdot \mathbf{C}$ denotes the dot product. %
In high-dimension spaces, (assuming $z$ and $\mathbf{C}$ are drawn from $\mathcal{N}(0, \sigma)$both the mean and variance of the distances scale linearly with dimension $n$:
\begin{align}
\mathbb{E}[|| z - \mathbf{C} ||_2^2] &= 2n\sigma^2 \\
\textbf{Var}[|| z - \mathbf{C} ||_2^2] &= 4n\sigma^4. 
\end{align}

By normalizing both $z$ and $\mathbf{C}$ with $\ell_2$ normalization (i.e., $||z||_2 = ||\mathbf{C}||_2 = 1$), the distance calculation simplifies to: 
\begin{equation} 
|| \ell_2(z) - \ell_2\mathbf{C}) ||_2^2 = 2(1 - \cos\theta)
\end{equation} where $\cos\theta$ represents the cosine similarity between $z$ and $\mathbf{C}$.

For $\ell_2$-normalized vectors, the expectation and variance of the squared Euclidean distance are as follows: 
\begin{align} 
\mathbb{E}[|| \ell_2(z) - \ell_2\mathbf{C}) ||_2^2] &= 2 \\
\textbf{Var}[|| \ell_2(z) - \ell_2\mathbf{C}) ||_2^2] &= \frac{4}{n-1} = \mathcal{O}(\frac{1}{n}). \end{align}

In high-dimensional spaces, most vectors in the codebook become nearly orthogonal to the query vector $z$. This results in similar distances from $z$ to most codebook vectors, converging towards $2$ as the dimension increases. %

However, the rate of this convergence is relatively slow. As dimensionality increases, the differences between the query vector and the vectors in the codebook become centralized around 2, with variance proportional to $\frac{1}{n}$. This highlights the inefficiency of directly quantifying high-dimensional vectors. Instead, quantifying individual components of high-dimensional vectors separately is more effective in preserving representational diversity and accuracy.

\subsection{Scaling Without Dimension Decomposition }
When dimension decomposition is not applied (i.e., GSQ with $G=1$), we explored the relationship between vocabulary size ($V$) and latent dimensionality ($D$) by tuning these parameters (\cref{fig:the_exp_scaling_latent_and_vocab}). The relationship between the rFID and the parameters $\log V$ and $D$ can be modeled as:

\begin{align} 
\text{rFID} &= \frac{B}{{\log V}^\alpha} + C \cdot D^\beta \\
&=  \frac{411.63}{({\log V})^{2.8375}} + 0.1601 \cdot D^{0.1956} \\
\end{align}

\newpage

\section{Ablation Studies of VAE}
\label{sec:appendix_exp_vae_setup}

\subsection{VAE Training configurations}
We list full training parameters here and \textcolor{red}{highlight} the optimized parameters that can improve the models' performance.
\begin{table*}[hbt!]
\centering
\begin{tabular}{ll}
\hline
\textbf{Parameter} & \textbf{Value} \\
\hline
\textbf{Training Parameters} & \\
Image Resolution & 128$\times$ 128 \\
Num Train Steps & 100,000 (20 epochs) \\
Gradient Clip & 2 \\
Mixed Precision & BF16 \\
Train Batch Size & 256 \\
Exponential Moving Average Beta & 0.999 \\
\hline
\textbf{Model Configuration} & \\
Down-sample-factor ($f$) & 8 \\
Hidden Channels & 128 \\
Channel Multipliers & [1, 2, 2, 4] \\
Encoder Layer Configs & [2, 2, 2, 2, 2] \\
Decoder Layer Configs & [2, 2, 2, 2, 2] \\
\hline
\textbf{Quantizer Settings} & \\
Embed Dimension ($D$) & 8 \\
Codebook Vocabulary ($V$) & 8192 \\
Group ($G$) & 1\\
Codebook Initialization & $\ell_2(\mathcal{N}(0, 1))$\\ 
Look-up Normalization & $\ell_2$ \\
\hline

\textbf{Loss weights} \\
Reconstruction Loss & 1.0 \\
Perceptual  Loss (LPIPS)& 1.0 \\
Commitment Loss & 0.25 \\

\hline
\textbf{VAE Optimizer} & \\
Base Learning Rate & $1 \times 10^{-4}$ \\
Learning Rate Scheduler & Fixed \\
Weight Decay & 0.05 \\
Betas & [0.9, 0.95]  \textcolor{red}{ $\rightarrow$ [0.9, 0.99]} \\
Epsilon & $1 \times 10^{-8}$ \\
\hline
\end{tabular}
\caption{\textbf{VAE-F8} Training Hyperparameters}
\label{tab:vae_hyperparameters}
\end{table*}

\subsection{Usage of Codebook Initialization Ablation Studies. }
\begin{figure}[hbt!]
    \centering
    \begin{minipage}[t]{0.6\textwidth}
        \centering
        \includegraphics[width=\textwidth]{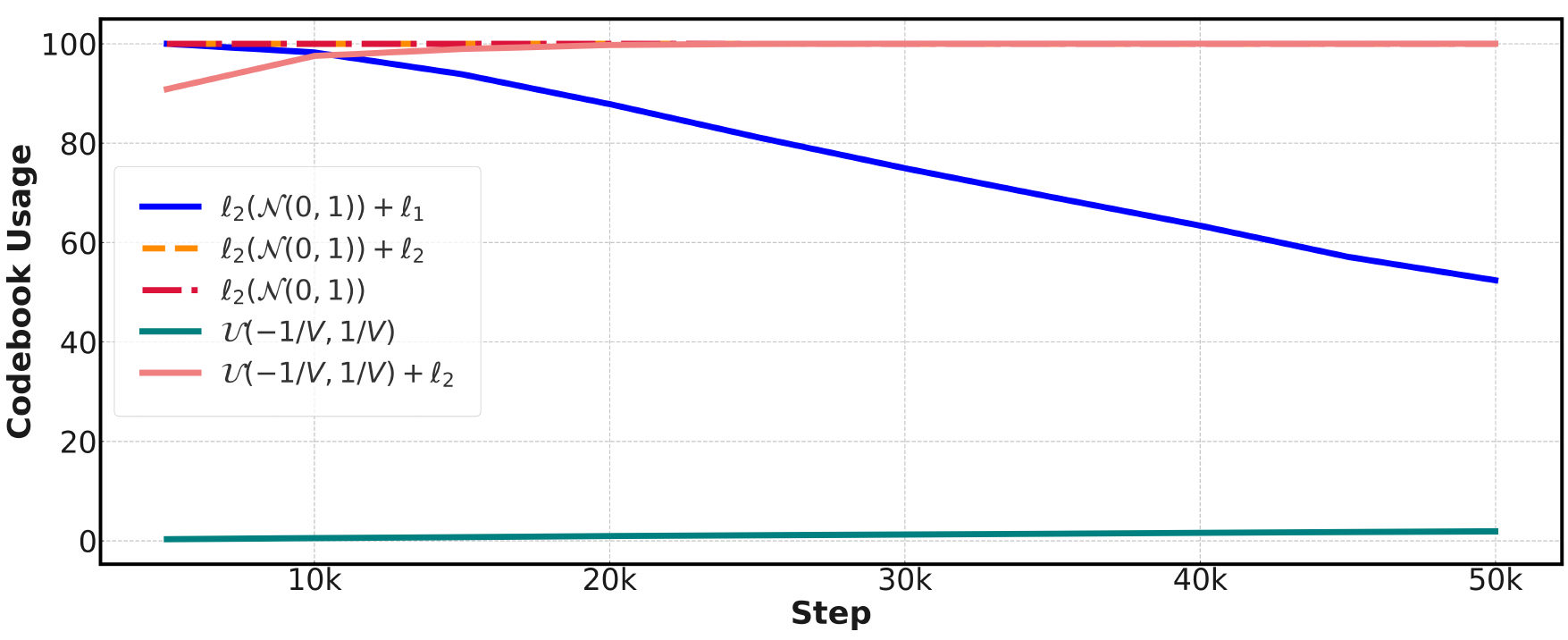}
    \end{minipage}
    \caption{Codebook usage during training for \textbf{GSQ-VAE-F8}. Our proposed $\ell_2(\mathcal{N}(0, 1))$ codebook initialisation, both with and without $\ell_2$, ensures consistent full codebook usage.}
    \label{fig:exp_codebook_init_codebook_usage}
\end{figure}

In \cref{sec:res_1.1}, we compared various codebook initialization methods and observed that $\ell_2$-normalized look-up in \textbf{VAE} achieves superior reconstruction performance and higher codebook usage. The detailed codebook usage during training is shown in \cref{fig:exp_codebook_init_codebook_usage}. Notably, the proposed spherical initialisation ensures 100\% codebook usage throughout the training process, unlike uniform initialisation.

\begin{figure}[hbt!]
    \centering
    \begin{minipage}[t]{0.6\textwidth}
        \centering
        \includegraphics[width=\textwidth]{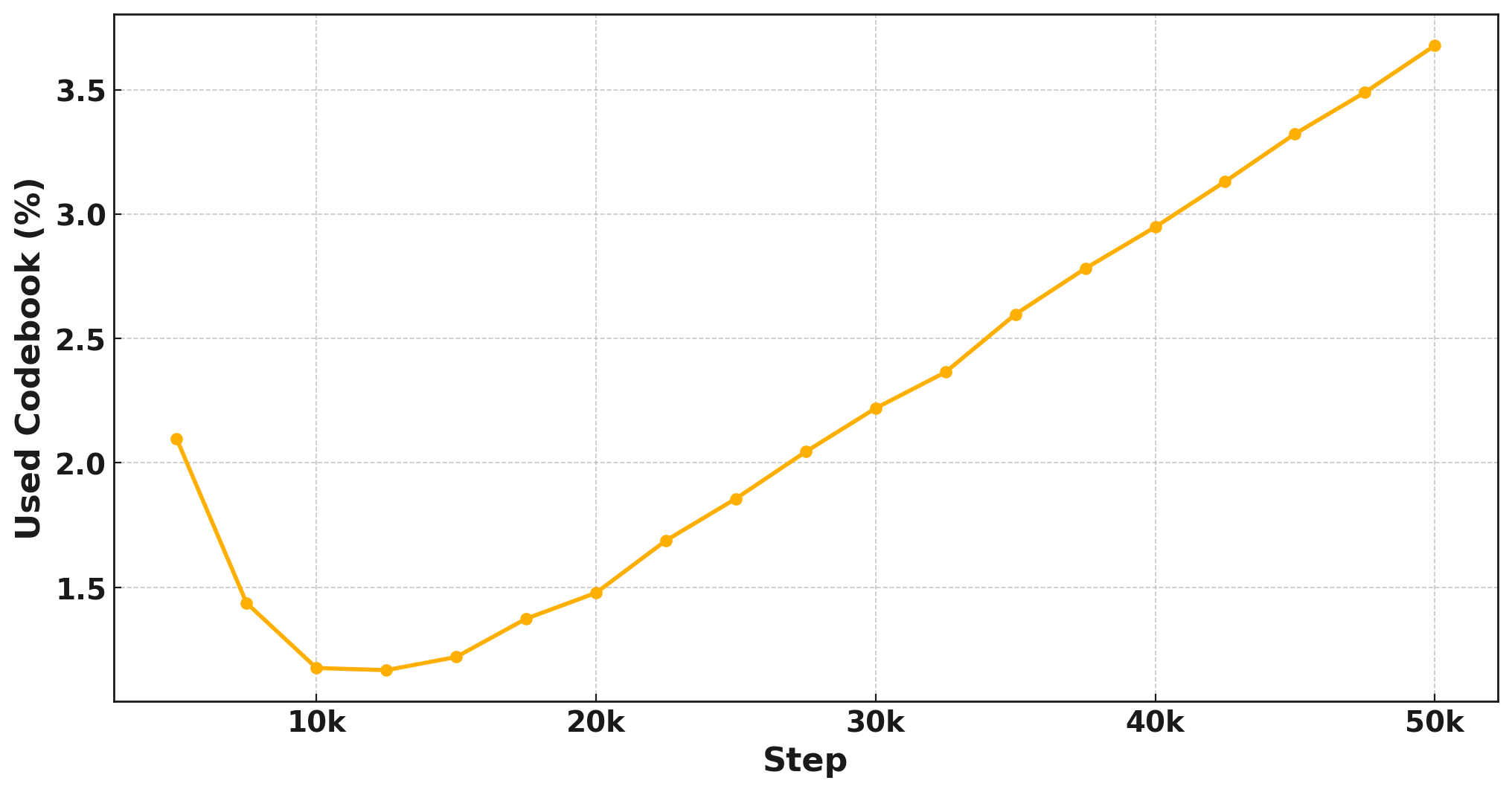}
    \end{minipage}
    \caption{Codebook usage for GSQ-GAN-F16-D16G4 training with a \textbf{uniformly} initialised codebook.}
    \label{fig:gsq-f16-g4-uniform}
\end{figure}

To further analyse the impact, we trained an additional model, GSQ-GAN-F16, with $G=4$ and a 256k vocabulary size, using a codebook initialised with a uniform distribution. As summarised in \cref{tab:exp_gvq_g_ablation}, the rFID of our proposed method is 0.52, while the uniform distribution case exhibits a degraded rFID of 0.66. More critically, the codebook usage drops significantly to just 3.68\% with uniform initialisation, as illustrated in \cref{fig:gsq-f16-g4-uniform}.

\newpage
\subsection{Learning Rate Scheduler}

\begin{figure}[hbt!]
    \centering
    \includegraphics[width=.6\textwidth]{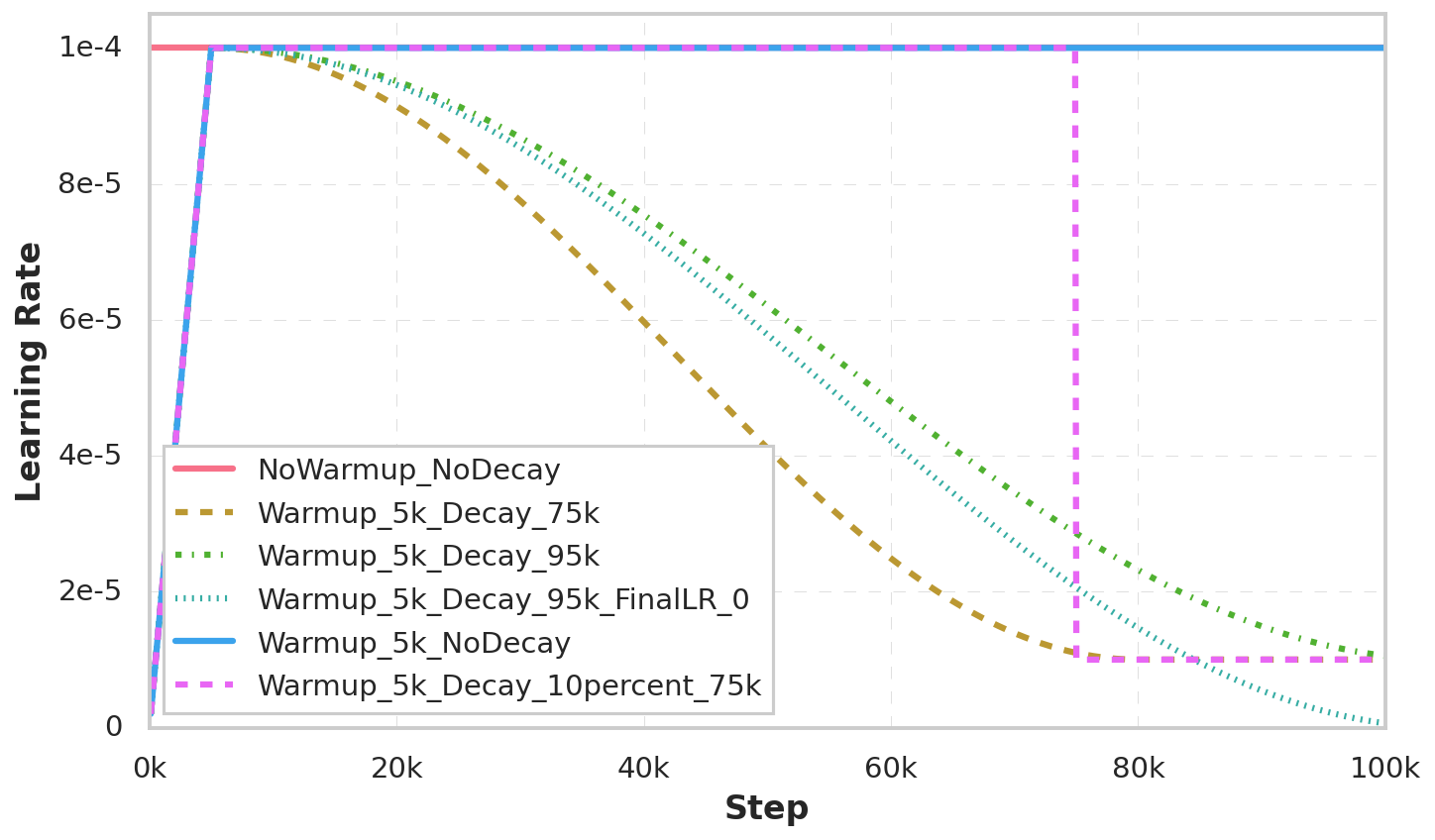}
    \caption{The learning rate schedules for \textbf{GSQ-VAE-F8} training. }
    \label{fig:exp_vae_lr_sched}
\end{figure}
In \cref{sec:res_1.4}, we compared five different learning rate schedulers against a constant learning rate for \textbf{GSQ-VAE-F8}. The detailed learning rate schedules relative to training steps are depicted in \cref{fig:exp_vae_lr_sched}.

\newpage
\subsection{VAE Reconstruction Visualization }
\begin{table}[hbt!]
\centering
\begin{tabular}{cccc}
\includegraphics[width=0.21\textwidth]{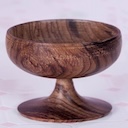} &
\includegraphics[width=0.21\textwidth]{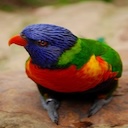} &
\includegraphics[width=0.21\textwidth]{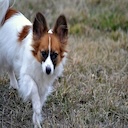} &
\includegraphics[width=0.21\textwidth]{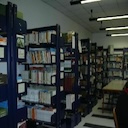} \\
\multicolumn{4}{c}{(a) Original images (128$\times$128 resolution)} \\
\includegraphics[width=0.21\textwidth]{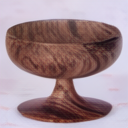} &
\includegraphics[width=0.21\textwidth]{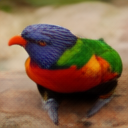} &
\includegraphics[width=0.21\textwidth]{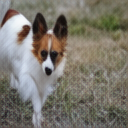} &
\includegraphics[width=0.21\textwidth]{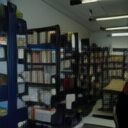} \\
\multicolumn{4}{c}{(b) Reconstruction results by \textbf{}{VAE-F8} } \\
\includegraphics[width=0.21\textwidth]{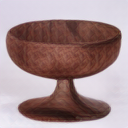} &
\includegraphics[width=0.21\textwidth]{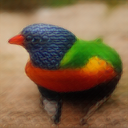} &
\includegraphics[width=0.21\textwidth]{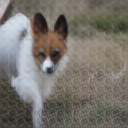} &
\includegraphics[width=0.21\textwidth]{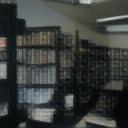} \\
\multicolumn{4}{c}{(c) With Depth2Scale} \\
\includegraphics[width=0.21\textwidth]{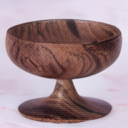} &
\includegraphics[width=0.21\textwidth]{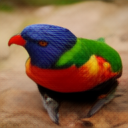} &
\includegraphics[width=0.21\textwidth]{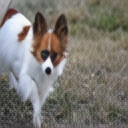} &
\includegraphics[width=0.21\textwidth]{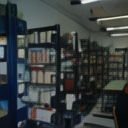} \\
\multicolumn{4}{c}{(d) With Adaptive Normalization} \\
\includegraphics[width=0.21\textwidth]{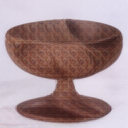} &
\includegraphics[width=0.21\textwidth]{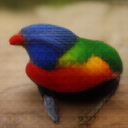} &
\includegraphics[width=0.21\textwidth]{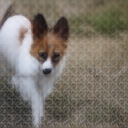} &
\includegraphics[width=0.21\textwidth]{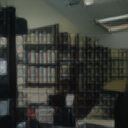} \\
\multicolumn{4}{c}{(e) With Depth2Scale and Adaptive Normalization} \\
\vspace{-0.5cm}
\end{tabular}
\caption{Reconstruction results of the \textbf{VAE-F8} model (in Section \ref{sec:res_1.2}) with ablation of Depth2Scale and Adaptive Normalization.}
\vspace{-0.5cm}
\end{table}

\newpage
\section{Ablation Studies of GAN}
\label{sec:appendix_exp_gan_setup}
\subsection{GAN Training configurations}
We list the full training parameters for \textbf{GAN} in \cref{tab:gan_hyperparameters}, with \textcolor{red}{highlighted} optimised parameters that significantly improve the model's performance. To achieve a high rFID without group decomposition, $\ell_2$ normalisation was omitted.

\begin{table*}[hbt!]
\centering
\begin{tabular}{ll}
\hline
\textbf{Parameter} & \textbf{Value} \\
\hline
\textbf{Training Parameters} & \\
Image Resolution & 128$\times$ 128 \\
Num Train Steps & 80,000 (16 epochs) \\
Gradient Clip & 2 \\
Mixed Precision & BF16 \\
Train Batch Size & 256 \\
Exponential Moving Average Beta & 0.999 \\
\hline
\textbf{Model Configuration} & \\
Down-sample-factor ($f$) & 8 \\
Hidden Channels & 128 \\
Channel Multipliers & [1, 2, 2, 4] \\
Encoder Layer Configs & [2, 2, 2, 2, 2] \\
Decoder Layer Configs & [2, 2, 2, 2, 2] \\
\hline
\textbf{Quantizer Settings} & \\
Embed Dimension ($D$) & 8 \\
Codebook Vocabulary ($V$) & 8192 \\
Group ($G$) & 1\\
Codebook Initialization & $\ell_2(\mathcal{N}(0, 1))$\\ 
Look-up Normalization &  \\
\hline
\textbf{Discriminator} & \\
Name & Dino Discriminator \\
Generator Loss & Non-Saturate \\
Discriminator Loss & Hinge \\
Dino-D Data Augmentation &  Cutout+Color+Translation \\
\hline
\textbf{Loss weights} \\
Reconstruction Loss & 1.0 \\
Perceptual  Loss (LPIPS)& 1.0 \\
Commitment Loss & 0.25 \\
Adversarial Loss & 0.1 \\
Discriminator Loss & 1.0 \\

\hline
\textbf{VAE Optimizer} & \\
Base Learning Rate & $1 \times 10^{-4}$ \textcolor{red}{ $\rightarrow$ $2 \times 10^{-4}$}\\
Learning Rate Scheduler & Fixed \\
Weight Decay & 0.05 \\
Betas & [0.9, 0.99]   \\
Epsilon & $1 \times 10^{-8}$ \\
\hline
\textbf{Discriminator Optimizer} & \\
Base Learning Rate & $1 \times 10^{-4}$ \textcolor{red}{ $\rightarrow$ $2 \times 10^{-4}$}\\
Learning Rate Scheduler & Fixed \\
Weight Decay & 0.05 \\
Betas & [0.5, 0.9] \textcolor{red}{ $\rightarrow$ [0.9, 0.99]} \\ 
Epsilon & $1 \times 10^{-8}$ \\
\hline
\end{tabular}
\caption{\textbf{GAN-F8} Training Hyperparameters}
\label{tab:gan_hyperparameters}
\end{table*}

\subsection{Discriminator Architecture}
We listed the network configurations of N-Layer, Dino and StyleGAN discriminators we used in GAN's ablation studies as follows:

\begin{table*}[hbt!]
\centering
\begin{tabular}{ll}
\hline
\textbf{Parameter} & \textbf{Value} \\
\hline
\textbf{N-Layer Discriminators (NLD)} & \\
Input Channels & 3 \\
Number of Channels & 64 \\
Number of Layers & 3 \\
\hline
\textbf{Style-GAN Discriminators (SGD)} & \\
Input Channels & 3 \\
Number of Channels & 128 \\
Channels Multiplier & [2, 4, 4, 4, 4] \\
\hline
\textbf{DINO Discriminators (DD)} & \\
Base Model & DinoV2\_vits14\_reg \\
Channels Multiplier & [2, 4, 4, 4, 4] \\
Features from layer & [2, 5, 8, 11] \\
\hline
\end{tabular}
\caption{Discriminator configurations}
\label{tab:discriminator_configs}
\end{table*}

\subsection{Adversarial and Discriminator Loss}
\label{sec:appendix_gan_loss}

We define the adversarial and discrimination loss as follows: the $\ell_{real}$ and $\ell_{fake}$ are logits of real and reconstructed images obtained by passing corresponding images to the discriminator.

\subsubsection*{Vanilla Discriminator Loss}
\begin{equation}
\mathcal{L}_{\text{vanilla\_discr}} = \frac{1}{2} \left( \mathbb{E}\left[\log(1 + e^{-\ell_{real}})\right] + \mathbb{E}\left[\log(1 + e^{\ell_{fake}})\right] \right)
\end{equation}

\subsubsection*{Vanilla Generator Loss}
\begin{equation}
\mathcal{L}_{\text{vanilla\_gen}} = \mathbb{E}\left[\log(1 + e^{-\ell_{fake}})\right]
\end{equation}

\subsubsection*{Hinge Generator Loss}
\begin{equation}
\mathcal{L}_{\text{hinge\_gen}} = -\mathbb{E}\left[\ell_{fake}\right]
\end{equation}

\subsubsection*{Hinge Discriminator Loss}
\begin{equation}
\mathcal{L}_{\text{hinge\_discr}} = \frac{1}{2} \left( \mathbb{E}\left[\max(0, 1 - \ell_{real})\right] + \mathbb{E}\left[\max(0, 1 + \ell_{fake})\right] \right)
\end{equation}

\subsubsection*{Non-Saturate Generator Loss}

\begin{equation}
\mathcal{L}_{\text{non\_saturate\_gen}} = \mathbb{E}\left[\text{ReLU}(\ell_{fake}) - \ell_{fake} \cdot 1 + \log\left(1 + e^{|\ell_{fake}|}\right)\right]
\end{equation}

\subsubsection*{Non-Saturate Discriminator Loss}

\begin{equation}
\mathcal{L}_{\text{real}} = \mathbb{E}\left[\text{ReLU}(\ell_{real}) - \ell_{real} \cdot 1 + \log\left(1 + e^{|\ell_{real}|}\right)\right]
\end{equation}

\begin{equation}
\mathcal{L}_{\text{fake}} = \mathbb{E}\left[\text{ReLU}(\ell_{fake}) - \ell_{fake} \cdot 0 + \log\left(1 + e^{|\ell_{fake}|}\right)\right]
\end{equation}

\begin{equation}
\mathcal{L}_{\text{non\_saturate\_discr}} = \frac{1}{2} \left( \mathcal{L}_{\text{real}} + \mathcal{L}_{\text{fake}} \right)
\end{equation}

\subsection{Failed Style-GAN Discriminator GAN's Training}
As discussed in the main paper, extensive ablations were conducted on Style-GAN Discriminator (\textit{SGD}) training. However, most experiments encountered numerical instability, resulting in $\mathit{NaN}$ errors. We provide a qualitative analysis of these failed runs by plotting training loss and evaluation rFID. We compare three combinations of discriminator losses: \textit{NV}, \textit{HH}, and \textit{NH}. These combinations were chosen based on their relatively better performance in the \textit{NLD} ablation studies (see \cref{tab:exp_gan_loss}).

\begin{figure}[hbt!]
    \centering
    \begin{minipage}[t]{0.55\textwidth}
        \centering
        \vspace{-0.2cm}
        \includegraphics[width=\textwidth]{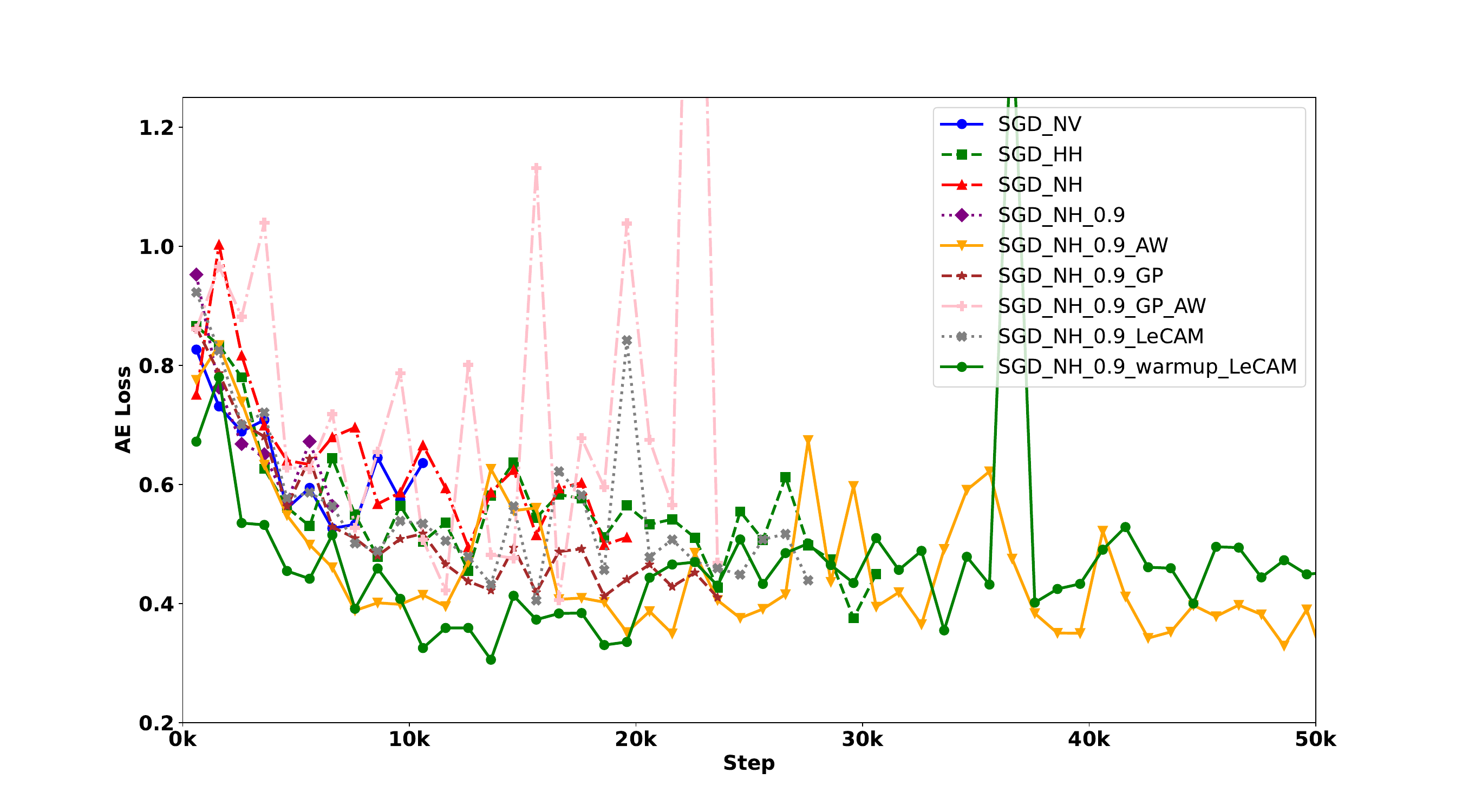}
        \subcaption{The summation of the generator (VAE) training loss of GSQ-GAN training with Style-GAN Discriminator.}
    \end{minipage}    
    \vspace{-0.15cm}
    \begin{minipage}[t]{0.55\textwidth}
    \centering
    \includegraphics[width=\textwidth]{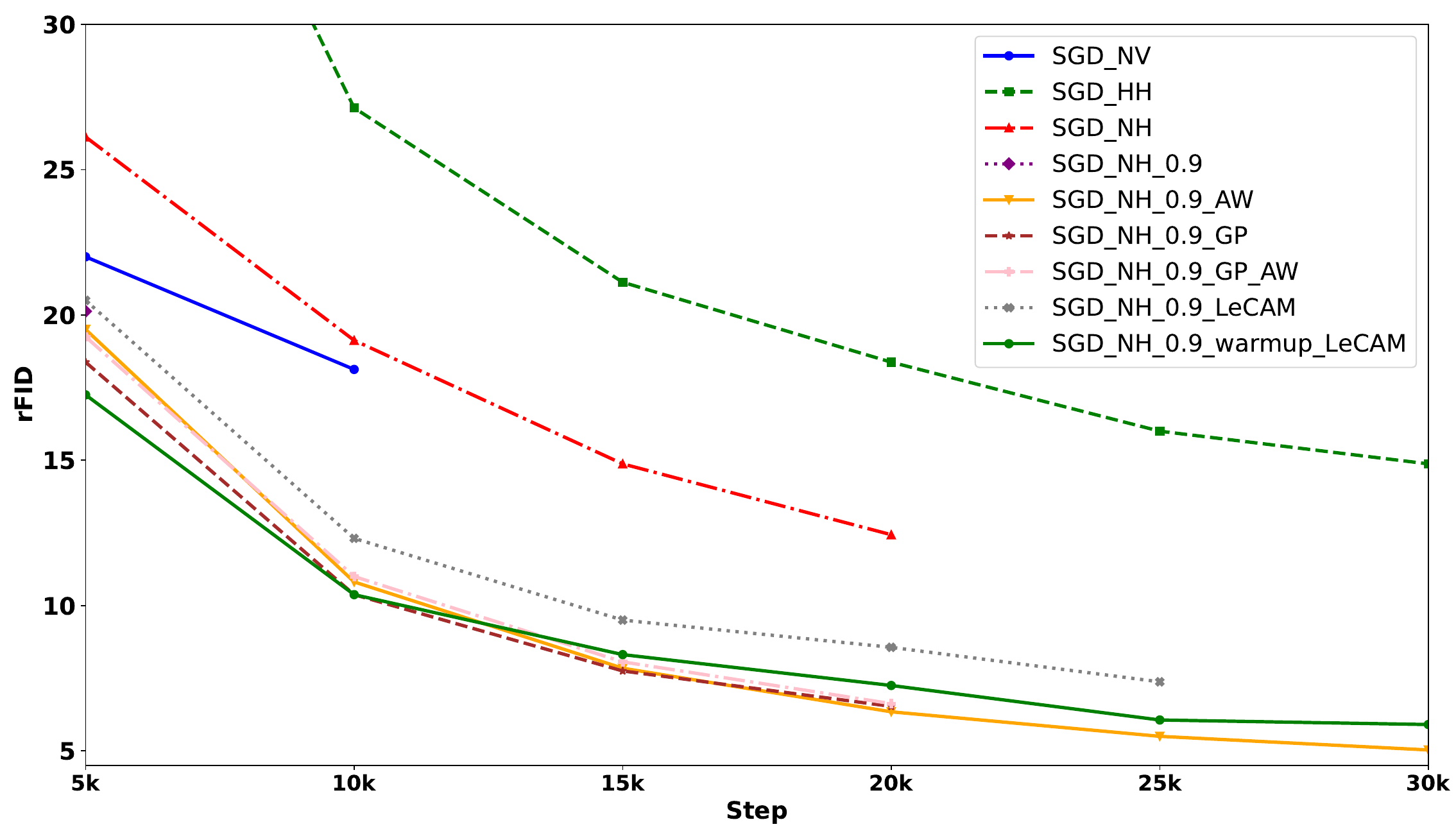}
    \subcaption{The RFID of GSQ-GAN trained with Style-GAN Discriminator and different combinations of discriminator loss and regularization.}
    \end{minipage}
    
    \caption{Style-GAN Discriminator training models' training loss and rFID are trained with different discriminator loss combinations and GAN regularization technologies.}
    \label{fig:sgd_failed_runs}
\end{figure}

As shown in \cref{fig:sgd_failed_runs}, training with \textit{NV} achieves the lowest rFID and exhibits more stable numerical behaviour than the other combinations. During this short training period, \textit{NV} performs better than \textit{NH}, achieving both lower rFID and lower training loss, consistent with the results of \textit{NLD}. However, \textit{SGD-NV} training fails abruptly at 10k steps due to $\mathit{NaN}$ errors. Training with \textit{NH} using the optimizer configuration $\beta=[0.9, 0.99]$ also fails before reaching the 10k step, previous studies (\textit{NLD} and \textit{DD}) suggesting that higher $\beta$ values boost model performance.

We further conducted ablation studies on GAN regularization techniques, including adaptive discriminator loss weights, LeCAM regularization, gradient penalty, and generator warmup. The results are presented in \cref{fig:sgd_failed_runs}. Training with gradient penalty regularization demonstrates a robust and stable dynamic, with the model's loss decreasing smoothly and achieving lower rFID than other methods. In contrast, training with LeCAM regularization shows significantly unstable behaviour, as reflected by sharp peaks in the loss curves.

Gradient penalty and adaptive Weights perform best for Style-GAN Discriminator training among all the regularization methods, but when these two work together, the training will be highly unstable. Meanwhile, due to the high parameter count and computational FLOPs of \textit{SGD}, gradient penalty regularization and adaptive weights become computationally expensive, requiring additional backward passes during training. Consequently, it makes \textit{SGD} an impractical choice for efficient GAN training.

\newpage

\subsection{GAN Reconstruction Visualization }
\begin{table}[hbt!]
\centering
\begin{tabular}{cccc}
\includegraphics[width=0.21\textwidth]{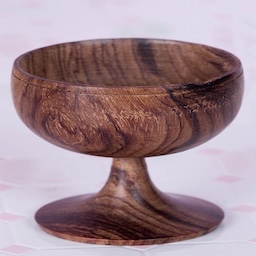} &
\includegraphics[width=0.21\textwidth]{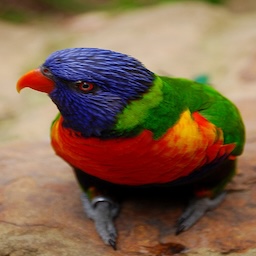} &
\includegraphics[width=0.21\textwidth]{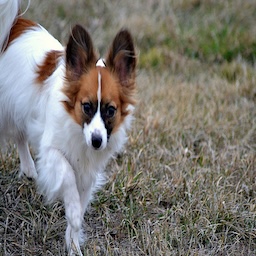} &
\includegraphics[width=0.21\textwidth]{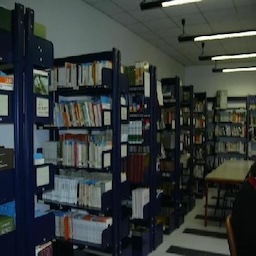} \\
\multicolumn{4}{c}{(a) Orignal images (128$\times$128 resolution)} \\
\includegraphics[width=0.21\textwidth]{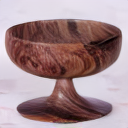} &
\includegraphics[width=0.21\textwidth]{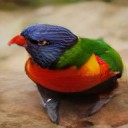} &
\includegraphics[width=0.21\textwidth]{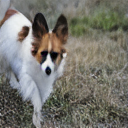} &
\includegraphics[width=0.21\textwidth]{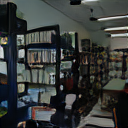} \\
\multicolumn{4}{c}{(b) Reconstruction results by with NLD-NV discriminators} \\
\includegraphics[width=0.21\textwidth]{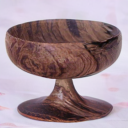} &
\includegraphics[width=0.21\textwidth]{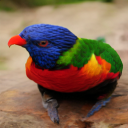} &
\includegraphics[width=0.21\textwidth]{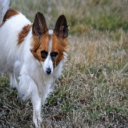} &
\includegraphics[width=0.21\textwidth]{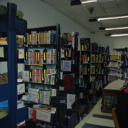} \\
\multicolumn{4}{c}{(c) Reconstruction results by with DD-NH discriminators} \\
\includegraphics[width=0.21\textwidth]{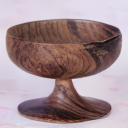} &
\includegraphics[width=0.21\textwidth]{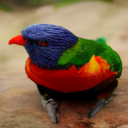} &
\includegraphics[width=0.21\textwidth]{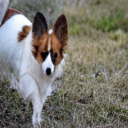} &
\includegraphics[width=0.21\textwidth]{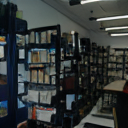} \\
\multicolumn{4}{c}{(d) Reconstruction results by with NLD-NV discriminators and $\beta=[0.9, 0.99]$} \\
\includegraphics[width=0.21\textwidth]{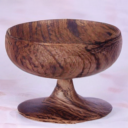} &
\includegraphics[width=0.21\textwidth]{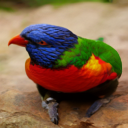} &
\includegraphics[width=0.21\textwidth]{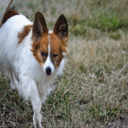} &
\includegraphics[width=0.21\textwidth]{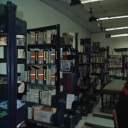} \\
\multicolumn{4}{c}{(e) Reconstruction results by with DD-NH discriminators and $\beta=[0.9, 0.99]$} \\
\vspace{-0.5cm}
\end{tabular}
\caption{Reconstruction results of the \textbf{GAN-F8} models (see Section \ref{sec:res_2.1}) , trained with different discriminators.}
\vspace{-0.5cm}
\end{table}

\newpage

\section{Scaling Behaviors}
\label{sec:appendix_scaling_latent}
This section details the training parameters for the \textbf{GAN} scaling experiments. All models were trained on the $256\times256$ resolution ImageNet dataset. Each scaling ablation study focuses on the latent dimension and codebook vocabulary size.
\begin{table*}[hbt!]
\centering
\begin{tabular}{ll}
\hline
\textbf{Parameter} & \textbf{Value} \\
\hline
\textbf{Training Parameters} & \\
Image Resolution & 256$\times$ 256 \\
Num Train Steps & 50,000 (20 epochs) \\
Gradient Clip & 2 \\
Mixed Precision & BF16 \\
Train Batch Size & 512 \\
Exponential Moving Average Beta & 0.999 \\
\hline
\textbf{Model Configuration} & \\
Down-sample-factor ($f$) & 8 \\
Hidden Channels & 128 \\
Channel Multipliers & [1, 2, 2, 4] \\
Encoder Layer Configs & [2, 2, 2, 2, 2] \\
Decoder Layer Configs & [2, 2, 2, 2, 2] \\
\hline
\textbf{Discriminator} & \\
Name & Dino Discriminator \\
Generator Loss & Non-Saturate \\
Discriminator Loss & Hinge \\
Dino-D Data Augmentation &  Cutout+Color+Translation \\
\hline
\textbf{Loss weights} \\
Reconstruction Loss & 1.0 \\
Perceptual  Loss (LPIPS)& 1.0 \\
Commitment Loss & 0.25 \\
Adversarial Loss & 0.1 \\
Discriminator Loss & 1.0 \\
\hline
\textbf{VAE and Discriminator  Optimizer} & \\
Base Learning Rate & $2 \times 10^{-4}$ \\
Learning Rate Scheduler & Fixed \\
Weight Decay & 0.05 \\
Betas & [0.9, 0.99]   \\
Epsilon & $1 \times 10^{-8}$ \\
\hline
\end{tabular}
\caption{\textbf{GAN-F8} Training Hyperparameters}
\label{tab:scaling_hyperparameters}
\end{table*}

In the network capacity scaling experiments described in \cref{sec:res_3.1}, the model names correspond to their respective \textit{Channel Multipliers}. The default \textit{depth} of the network is set to two for each block (\textit{Encoder Layer Configs} and \textit{Decoder Layer Configs}). For the \textbf{Deeper} network configuration, the \textit{Encoder Layer Configs} are set to $[4,3,4,3,4,4]$ and the \textit{Decoder Layer Configs} to $[3,4,3,4,4,4]$, following the architectural design principles outlined in MagVit2 \cite{magvit2}.

\newpage
\begin{table}[H]
\centering
\begin{tabular}{cccc}
\includegraphics[width=0.21\textwidth]{figures_results/original/n03443371_goblet.jpeg} &
\includegraphics[width=0.21\textwidth]{figures_results/original/n01820546_lorikeet.jpeg} &
\includegraphics[width=0.21\textwidth]{figures_results/original/n02086910_papillon.jpeg} &
\includegraphics[width=0.21\textwidth]{figures_results/original/n03661043_library.jpeg} \\
\multicolumn{4}{c}{(a) Orignal images (256$\times$256 resolution) } \\
\includegraphics[width=0.21\textwidth]{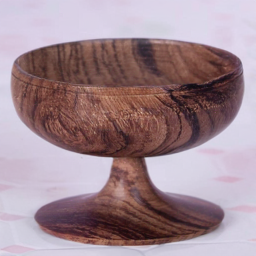} &
\includegraphics[width=0.21\textwidth]{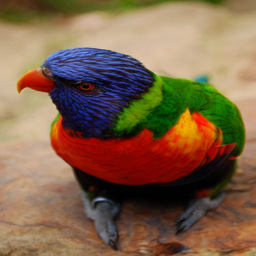} &
\includegraphics[width=0.21\textwidth]{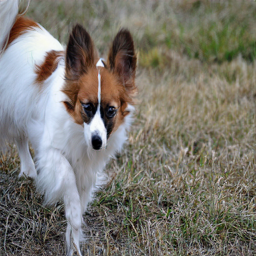} &
\includegraphics[width=0.21\textwidth]{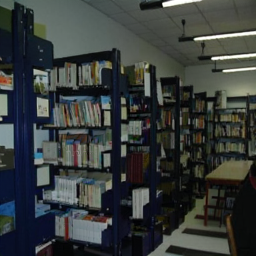} \\
\multicolumn{4}{c}{(b) GSQ-GAN-F8, $D=16$, $G=1$} \\
\includegraphics[width=0.21\textwidth]{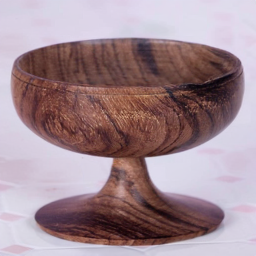} &
\includegraphics[width=0.21\textwidth]{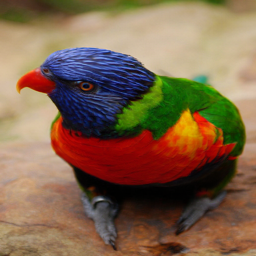} &
\includegraphics[width=0.21\textwidth]{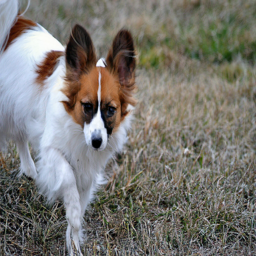} &
\includegraphics[width=0.21\textwidth]{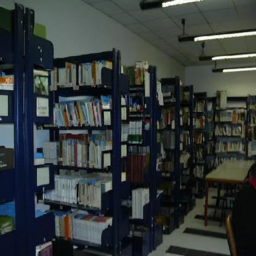} \\
\multicolumn{4}{c}{(b) GSQ-GAN-F8, $D=32$, $G=2$} \\
\includegraphics[width=0.21\textwidth]{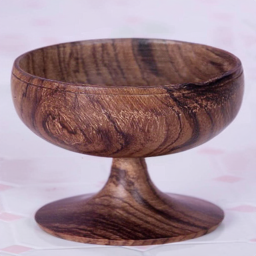} &
\includegraphics[width=0.21\textwidth]{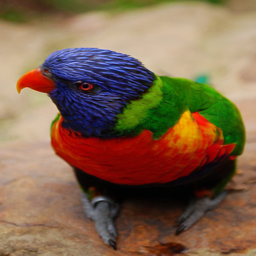} &
\includegraphics[width=0.21\textwidth]{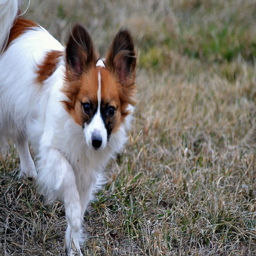} &
\includegraphics[width=0.21\textwidth]{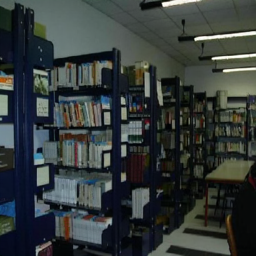} \\
\multicolumn{4}{c}{(b) GSQ-GAN-F8, $D=64$, $G=4$} \\
\vspace{-0.5cm}
\end{tabular}
\caption{Scaling latent dimension for \textbf{GSQ-GAN-F8} model. The models are detailed in \cref{fig:exp_scaling_factor}.}
\vspace{-0.5cm}
\end{table}

\begin{table}[H]
\centering
\begin{tabular}{cccc}
\includegraphics[width=0.21\textwidth]{figures_results/original/n03443371_goblet.jpeg} &
\includegraphics[width=0.21\textwidth]{figures_results/original/n01820546_lorikeet.jpeg} &
\includegraphics[width=0.21\textwidth]{figures_results/original/n02086910_papillon.jpeg} &
\includegraphics[width=0.21\textwidth]{figures_results/original/n03661043_library.jpeg} \\
\multicolumn{4}{c}{(a) Orignal images (256$\times$256 resolution) } \\
\includegraphics[width=0.21\textwidth]{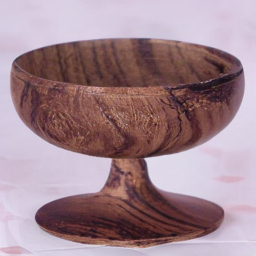} &
\includegraphics[width=0.21\textwidth]{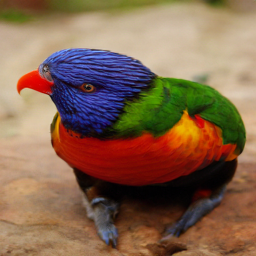} &
\includegraphics[width=0.21\textwidth]{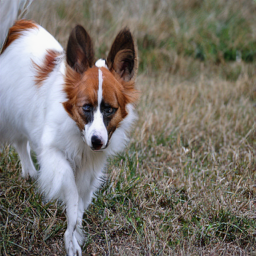} &
\includegraphics[width=0.21\textwidth]{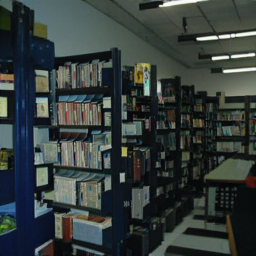} \\
\multicolumn{4}{c}{(b) GSQ-GAN-F16, $D=16$, $G=1$} \\
\includegraphics[width=0.21\textwidth]{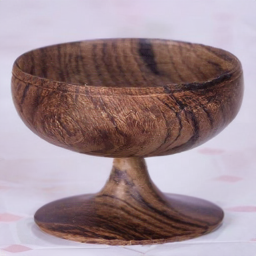} &
\includegraphics[width=0.21\textwidth]{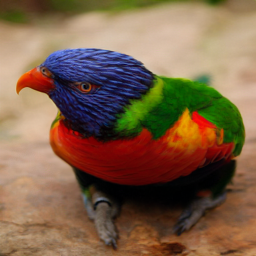} &
\includegraphics[width=0.21\textwidth]{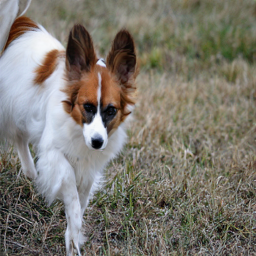} &
\includegraphics[width=0.21\textwidth]{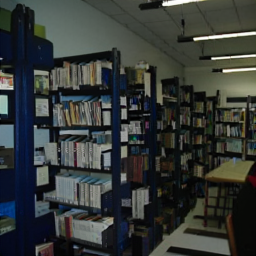} \\
\multicolumn{4}{c}{(c) GSQ-GAN-F16, $D=32$, $G=2$} \\
\includegraphics[width=0.21\textwidth]{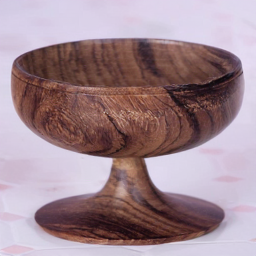} &
\includegraphics[width=0.21\textwidth]{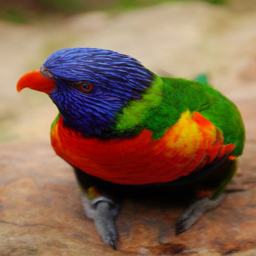} &
\includegraphics[width=0.21\textwidth]{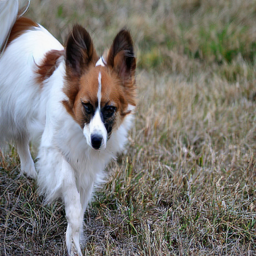} &
\includegraphics[width=0.21\textwidth]{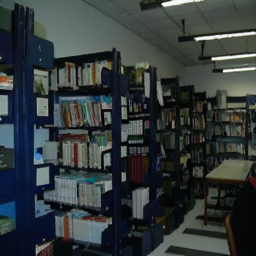} \\
\multicolumn{4}{c}{(c) GSQ-GAN-F16, $D=64$, $G=4$} \\
\vspace{-0.5cm}
\end{tabular}
\caption{Scaling latent dimension for \textbf{GSQ-GAN-F16} model. The models are detailed in \cref{fig:exp_scaling_factor}.}
\vspace{-0.5cm}
\end{table}

\begin{table}[H]
\centering
\begin{tabular}{cccc}
\includegraphics[width=0.21\textwidth]{figures_results/original/n03443371_goblet.jpeg} &
\includegraphics[width=0.21\textwidth]{figures_results/original/n01820546_lorikeet.jpeg} &
\includegraphics[width=0.21\textwidth]{figures_results/original/n02086910_papillon.jpeg} &
\includegraphics[width=0.21\textwidth]{figures_results/original/n03661043_library.jpeg} \\
\multicolumn{4}{c}{(a) Orignal images (256$\times$256 resolution) } \\
\includegraphics[width=0.21\textwidth]{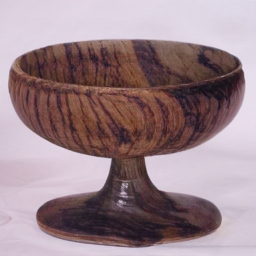} &
\includegraphics[width=0.21\textwidth]{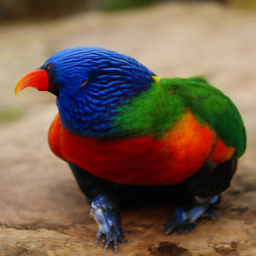} &
\includegraphics[width=0.21\textwidth]{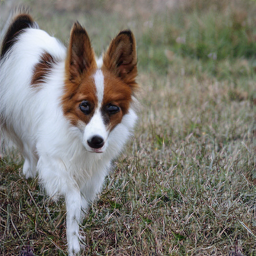} &
\includegraphics[width=0.21\textwidth]{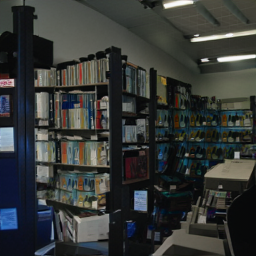} \\
\multicolumn{4}{c}{(b) GSQ-GAN-F32, $D=16$, $G=1$} \\
\includegraphics[width=0.21\textwidth]{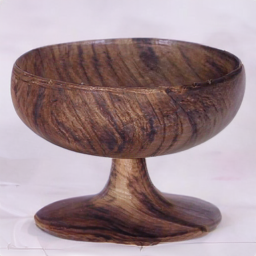} &
\includegraphics[width=0.21\textwidth]{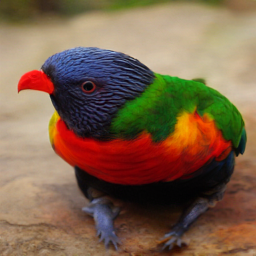} &
\includegraphics[width=0.21\textwidth]{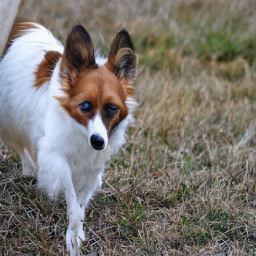} &
\includegraphics[width=0.21\textwidth]{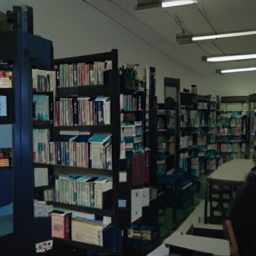} \\
\multicolumn{4}{c}{(c) GSQ-GAN-F32, $D=32$, $G=2$} \\
\includegraphics[width=0.21\textwidth]{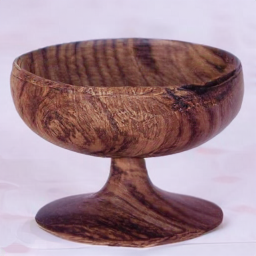} &
\includegraphics[width=0.21\textwidth]{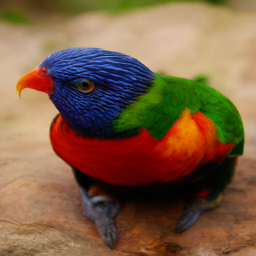} &
\includegraphics[width=0.21\textwidth]{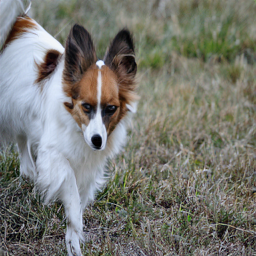} &
\includegraphics[width=0.21\textwidth]{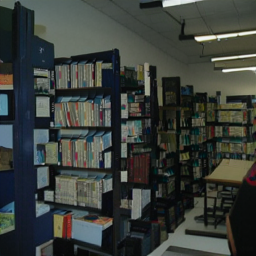} \\
\multicolumn{4}{c}{(d) GSQ-GAN-F32, $D=64$, $G=4$} \\
\includegraphics[width=0.21\textwidth]{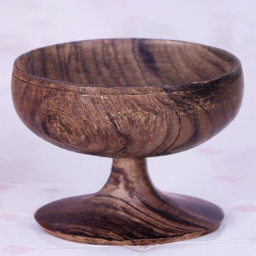} &
\includegraphics[width=0.21\textwidth]{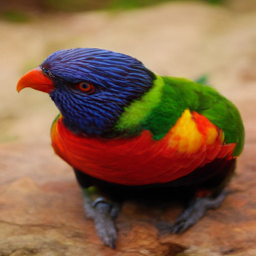} &
\includegraphics[width=0.21\textwidth]{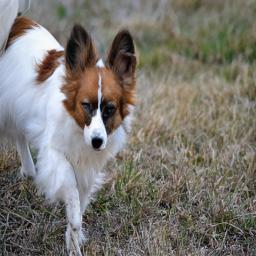} &
\includegraphics[width=0.21\textwidth]{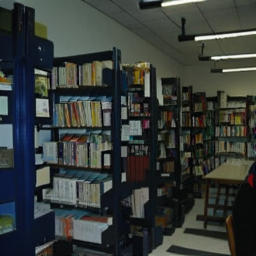} \\
\multicolumn{4}{c}{(e) GSQ-GAN-F32, $D=128$, $G=8$} \\
\vspace{-0.5cm}
\end{tabular}
\caption{Scaling latent dimension for \textbf{GSQ-GAN-F32} model. The models are detailed in \cref{fig:exp_scaling_factor}.}
\vspace{-0.5cm}
\end{table}

\begin{table}[H]
\centering
\begin{tabular}{cccc}
\includegraphics[width=0.21\textwidth]{figures_results/original/n03443371_goblet.jpeg} &
\includegraphics[width=0.21\textwidth]{figures_results/original/n01820546_lorikeet.jpeg} &
\includegraphics[width=0.21\textwidth]{figures_results/original/n02086910_papillon.jpeg} &
\includegraphics[width=0.21\textwidth]{figures_results/original/n03661043_library.jpeg} \\
\multicolumn{4}{c}{(a) Orignal images (256$\times$256 resolution) } \\
\includegraphics[width=0.21\textwidth]{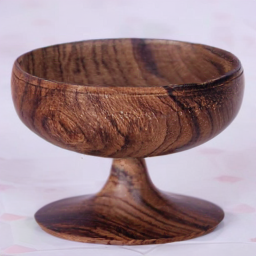} &
\includegraphics[width=0.21\textwidth]{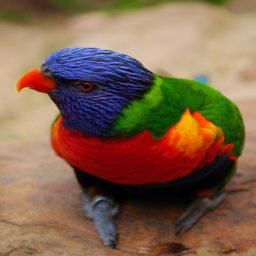} &
\includegraphics[width=0.21\textwidth]{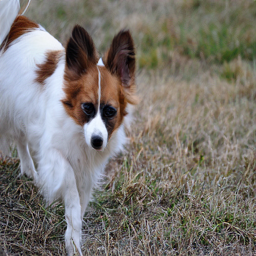} &
\includegraphics[width=0.21\textwidth]{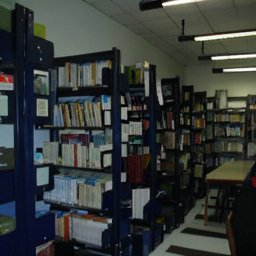} \\
\multicolumn{4}{c}{(b) GSQ-GAN-F8, $D=64$, $G=1$} \\
\includegraphics[width=0.21\textwidth]{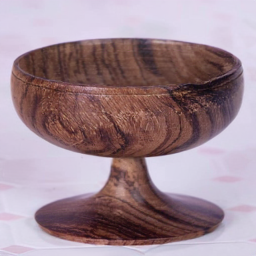} &
\includegraphics[width=0.21\textwidth]{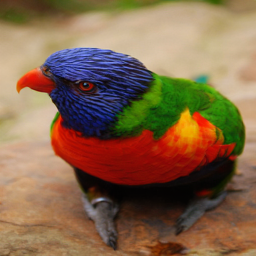} &
\includegraphics[width=0.21\textwidth]{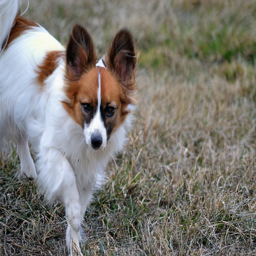} &
\includegraphics[width=0.21\textwidth]{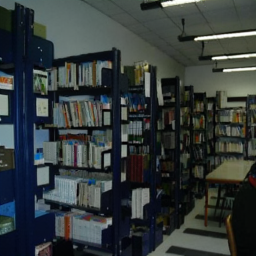} \\
\multicolumn{4}{c}{(c) GSQ-GAN-F8, $D=64$, $G=2$} \\
\includegraphics[width=0.21\textwidth]{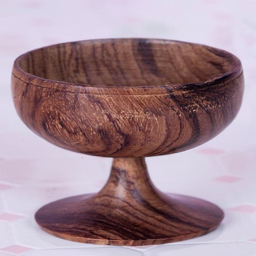} &
\includegraphics[width=0.21\textwidth]{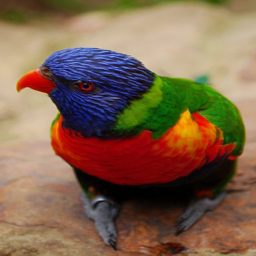} &
\includegraphics[width=0.21\textwidth]{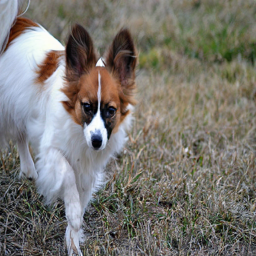} &
\includegraphics[width=0.21\textwidth]{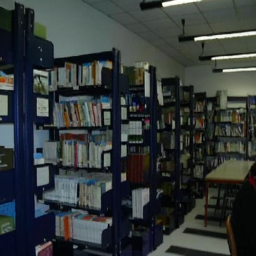} \\
\multicolumn{4}{c}{(c) GSQ-GAN-F8, $D=64$, $G=4$} \\
\includegraphics[width=0.21\textwidth]{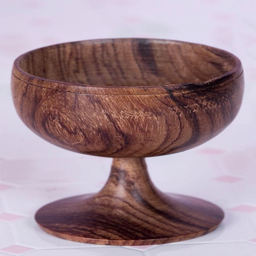} &
\includegraphics[width=0.21\textwidth]{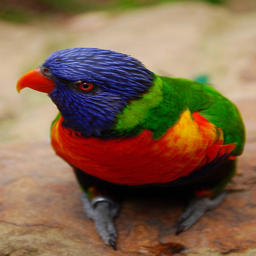} &
\includegraphics[width=0.21\textwidth]{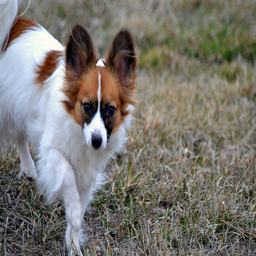} &
\includegraphics[width=0.21\textwidth]{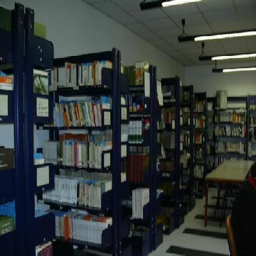} \\
\multicolumn{4}{c}{(c) GSQ-GAN-F8, $D=64$, $G=16$} \\
\vspace{-0.5cm}
\end{tabular}
\caption{Scaling latent dimension for \textbf{GSQ-GAN-F8-D64} model. The models are detailed in Section \ref{sec:res_3.3}.}
\vspace{-0.5cm}
\end{table}

\end{document}